\documentclass{article}

     \usepackage[nonatbib,preprint]{neurips_2018}

\usepackage[utf8]{inputenc} % allow utf-8 input
\usepackage[T1]{fontenc}    % use 8-bit T1 fonts

\usepackage[hidelinks]{hyperref}
\usepackage[nameinlink]{cleveref}
\usepackage[numbers]{natbib}

\usepackage{url}            % simple URL typesetting
\usepackage{booktabs}       % professional-quality tables
\usepackage{amsfonts}       % blackboard math symbols
\usepackage{nicefrac}       % compact symbols for 1/2, etc.
\usepackage{microtype}      % microtypography
\usepackage{amsmath}

\DeclareMathOperator{\E}{\mathbb{E}}
\DeclareMathOperator{\Pro}{\mathbb{P}}
\usepackage{amssymb}
\usepackage{mathtools}
\usepackage{esvect}
\usepackage{algorithm}% http://ctan.org/pkg/algorithm
\usepackage{algpseudocode}% http://ctan.org/pkg/algorithmicx
\usepackage{amsmath}
\usepackage{bm}
\usepackage{amsthm}
\usepackage{physics}
\usepackage{verbatim}
\usepackage{subfig}
\usepackage{graphicx}
\usepackage{csquotes}
\usepackage{natbib}
\newtheorem*{theorem1*}{Theorem 0}
\newtheorem*{theorem2*}{Theorem 1}
\newtheorem*{theorem11*}{Theorem 2}
\newtheorem*{theorem15*}{Theorem 3}
\newtheorem*{theorem19*}{Theorem 4}
\newtheorem*{theorem20*}{Theorem 5}
\newtheorem*{theorem3*}{Lemma 1}
\newtheorem*{theorem4*}{Lemma 2}
\newtheorem*{theorem5*}{Lemma 3}
\newtheorem*{theorem6*}{Corollary 1}
\newtheorem*{theorem8*}{Corollary 2}
\newtheorem*{theorem16*}{Corollary 3}
\newtheorem*{theorem18*}{Corollary 4}
\newtheorem*{theorem9*}{Lemma 3 (Informal)}
\newtheorem*{theorem10*}{Lemma 5}
\newtheorem*{theorem12*}{Assumption 1}
\newtheorem*{theorem14*}{Assumption 0}
\usepackage{cleveref}

\title{On the Separability of Classes with the Cross-Entropy Loss Function}

\author{%
  Rudrajit Das\\ %, Subhasis Chaudhuri \\
  Department of Electrical Engineering\\
  Indian Institute of Technology Bombay\\
  \texttt{rudrajit1503@gmail.com}\\ %, \texttt{sc@ee.iitb.ac.in}\\
   \And
   Subhasis Chaudhuri \\
   Department of Electrical Engineering \\
   Indian Institute of Technology Bombay \\
   \texttt{sc@ee.iitb.ac.in} \\
}
\begin{document}

\maketitle

\begin{abstract}
 In this paper, we focus on the separability of classes with the cross-entropy loss function for classification problems by theoretically analyzing the intra-class distance and inter-class distance 
 (i.e. the distance between any two points belonging to the same class and different classes, respectively) 
 in the feature space, i.e. the space of representations learnt by neural networks. Specifically, we consider an arbitrary %neural 
 network architecture having a fully connected final layer with Softmax activation and trained using the cross-entropy loss. 
 We derive expressions for the value and the 
 distribution of the squared $\ell_{2}$ norm of the product of a network dependent matrix and a random intra-class and inter-class distance vector (i.e. the vector between any two points belonging to the same class and different classes), respectively, in the
 learnt 
 feature space %learnt by neural networks 
 (or the transformation of the original data) just before Softmax activation, as a function of the cross-entropy loss value. 
 The main result of our analysis is the derivation of a lower bound for the probability with which the inter-class distance is more than the intra-class distance in this %transformed 
 feature 
 space, as a function of the loss value. 
 We do so by leveraging some empirical statistical observations with %reasonable
 mild assumptions and sound theoretical analysis. 
 {As per intuition, the probability with which the inter-class distance is more than the intra-class distance decreases as the loss value increases, i.e. the classes are better separated when the loss value is low.}
 To the best of our knowledge, this is the first work of theoretical nature trying to explain the separability of classes in the  
 feature 
 %representation 
 space learnt by neural networks trained with the cross-entropy loss function.
\end{abstract}

\section{Introduction}
\label{sec:0}

%\textbf{FILL IT!}
%\medskip{}
Classification problems are ubiquitous in machine learning. 
Deep neural networks (\cite{krizhevsky2012imagenet,simonyan2014very,he2016identity,szegedy2017inception}) have been immensely successful in solving supervised classification problems. A central part of these networks is the final Softmax layer %which is used 
to obtain the 
%predictions
predicted probabilities of belonging to each class. 
%and the use of the cross-entropy loss function as our objective function %(which is minimized in order to obtain the optimal parameter values of the network)
The most commonly used loss function %to determine the optimal network parameter values is the cross-entropy loss function. 
for classification problems is the cross-entropy loss. %function. 
The cross-entropy loss along with the final Softmax layer try to obtain class-wise linear partitions of the representation/transformation of the original data just before the Softmax layer by maximizing the likelihood of the data with respect to the network parameters. 

On the other hand, more conventional techniques such as Fisher Linear Discriminant Analysis (LDA) try to obtain a linear separation of the data by maximizing the separation between the class-means (relative to the sum of the variances of the data %assigned to
in each class). 
Similarly, in an SVM (\cite{cortes1995support}) maximum margin (binary) classifier, the goal is to find a linear separation of the data (either in its original space or after using a kernel) by obtaining two parallel hyper-planes that partition the two classes of data such that the distance between them (which is basically the "margin") is as large as possible.
Since the cross-entropy loss function is based on the maximum likelihood estimate approach, it does not %\textbf{overtly/directly} 
directly 
try to maximize the separation between the classes, which LDA and SVM do. 

Two related self-explanatory terms used in the literature are "intra-class compactness" and "inter-class separability". There is no shortage of works (\cite{luo2019mathcal,liu2016large,liu2017sphereface,chen2018virtual,gao2018margin,sun2014deep,zhou2019separability}) which point out %empirically 
that the vanilla cross-entropy loss with Softmax does not quite promote high intra-class compactness and inter-class separability and therefore propose 
%novel 
modified loss functions/architectures which address this issue. However, there is not much of theoretical justification %/motivation 
to %sufficiently 
properly 
explain this issue for the cross-entropy loss.

%In this work, we try to probabilistically show the separability of classes obtained using the cross-entropy loss function. Specifically, our main contribution is deriving a lower bound for the probability with which the inter-class distance (i.e. distance between any two random data points having different ground truth classes) is more than the intra-class distance (i.e. distance between any two random data points having the same ground truth class) in the feature space learnt by an arbitrary neural network (just before the final Softmax layer) as a function of the cross-entropy loss value. Thereby, we provide a mathematical quantification of the separability of classes attained in the feature space learnt by a network with the cross-entropy loss function.

In this work, we provide a %mathematical 
probabilistic
quantification of the separability of classes attained in the feature space learnt by a network trained with the cross-entropy loss, as a function of the loss value. Our main contributions are as follows. Firstly in \textbf{Theorem 1},
%In doing so, firstly, 
we provide an expression for the value and the complementary cumulative distribution function (ccdf) of the squared $\ell_{2}$ norm of the product of a network dependent matrix and a random intra-class distance vector (i.e. the vector between any two random data points having the same ground truth class) 
%(as explained in the abstract)
in %the transformed space. 
the feature space.
%the feature space learnt by a neural network. 
Secondly in \textbf{Theorem 2}, we provide a lower bound for the value as well as the ccdf of the squared $\ell_{2}$ norm of the product of a network dependent matrix and a random inter-class distance vector between any two classes (i.e. the vector between any two data points having different ground truth classes)
%(as explained in the abstract)
in the %transformed 
feature 
space. The network dependent matrix is the same for both the aforementioned cases if we consider the intra-class distance in a class say, $c$, and the inter-class distance between the same class $c$ and another class $c' \ne c$.
Thereafter in \textbf{Theorem 3}, considering two random points belonging to $c$ and one random point belonging to $c'$, we derive a lower bound on the probability with which the squared $\ell_{2}$ norm of the product of the previously mentioned network dependent matrix and the inter-class distance vector between the first point belonging to $c$ and the one belonging to $c'$ is more than the squared $\ell_{2}$ norm of the product of the same matrix and the intra-class distance vector between the two points belonging to $c$, times a certain factor (of our choice) %greater than 
$>1$. 
%Finally 
Next, in \textbf{Theorem 4} (%this is 
our main result), with some assumptions on the distribution of the entries of the aforementioned matrix, we derive a lower bound on the probability with which the inter-class distance (between the first point in $c$ and the one in $c'$) is more than the intra-class distance (between the two points in $c$) times a certain factor (of our choice) $\geq$ 1.
%Finally, with some assumptions on the distribution of the entries of the aforementioned matrix, we derive a lower bound on the probability with which the inter-class distance (between the first point belonging to $c$ and the one belonging to $c'$) is more than the intra-class distance (between the two points belonging to $c$).
Finally, in \textbf{Theorem 5}, we provide an expression for the expected per-class accuracy of the network model in consideration, as a function of the loss value, to relate accuracy with class separability.

To the best of our knowledge, this is the first theoretical attempt at quantifying the separability of classes, %in the feature space learnt by neural networks trained 
attained
with the cross-entropy loss function. %(along with Softmax). 
%We hope that
%In theory, our techniques can be extended to do the same task for other loss functions too.

\section{Problem Description and Preliminaries}
\label{sec:1}
Consider a classification problem with $C$ classes, labelled from $0$ through to $(C-1)$. Assume that we have trained a neural network architecture (be it a fully connected network or a convolutional network) having a fully connected final layer with the Softmax activation and using the cross-entropy loss (which is the most common model for classification problems) for the problem in hand. Denote this network by $\mathcal{N}_{c}$. 
Consider $m$ data points $\bm{x_{1}},\bm{x_{2}},\ldots,\bm{x_{m}}$ (which could be 1-D inputs such as vectors or 2-D inputs such as images etc.) whose ground truth classes are $c(\bm{x_{1}}),c(\bm{x_{2}}),\ldots,c(\bm{x_{m}})$, respectively. Let $y_{i}^{(j)} = \delta(j - c(\bm{x_{i}}))$, where $\delta(z) = 1$ if $z = 0$ and $\delta(z) = 0$ if $z \ne 0$. Finally, let the probability of $\bm{x_{i}}$ belonging to class $j$ predicted by our network $\mathcal{N}_{c}$ be denoted by $\widehat{y_{i}}^{(j)}$.
Then the cross-entropy loss is given as follows:
\small
\begin{equation}
\label{eq:1}
L = -\frac{1}{m}\sum_{i=1}^{m}\sum_{j=1}^{C}y_{i}^{(j)}\log(\widehat{y_{i}}^{(j)})
\end{equation}
\normalsize
Let us denote the transformed points (which are the features or representations learnt by the network) acting as inputs to the final Softmax layer, by $\phi(\bm{x})$. These are 1-D vectors of dimension $n$. Thus, $\bm{x_{1}},\bm{x_{2}},\ldots,\bm{x_{m}}$ get transformed to $\phi(\bm{x_{1}}),\phi(\bm{x_{2}}),\ldots,\phi(\bm{x_{m}})$, respectively. Denote the weights and biases of the last layer by $\bm{A}$ (which is a $C \times n$ matrix) and $\bm{b}$ (which is a $C \times 1$ vector). Let the $j^{\text{th}}$ ($1 \leq j \leq C$) row of $\bm{A}$ be denoted by $\bm{a_{j}}$. Similarly, let the $j^{\text{th}}$ ($1 \leq j \leq C$) element of $\bm{b}$ be denoted by ${b_{j}}$. Then we have:
\small
\begin{equation}
\label{eq:2}
%\widehat{y_{i}}^{(j)} = \frac{\exp(\bm{a_{j}}^{T}\phi(\bm{x_{i}}) + b_{j})}{\sum_{k=1}^{C} \exp(\bm{a_{k}}^{T}\phi(\bm{x_{i}}) + b_{k})}
\widehat{y_{i}}^{(j)} = {\exp(\bm{a_{j}}^{T}\phi(\bm{x_{i}}) + b_{j})}/{\sum_{k=1}^{C} \exp(\bm{a_{k}}^{T}\phi(\bm{x_{i}}) + b_{k})}
\end{equation}
\normalsize
%\medskip{}
%\begin{theorem14*}
%$\Big(-\log(\widehat{y_{i}}^{(c(\bm{x_{i}}))})\Big)^{1/\beta}$ approximately follows an exponential distribution (say with mean $\mu$) where $\beta$ is a constant $> 1$.
%\end{theorem14*}
%\medskip{}
Empirically, we observed that $\Big(-\log(\widehat{y_{i}}^{(c(\bm{x_{i}}))})\Big)^{1/\beta}$ approximately follows an exponential distribution (say with mean $\mu$) where $\beta$ is a constant depending on the network and dataset properties. %$> 1$. 
For MNIST and CIFAR-10, $\beta \approx 4$ on the networks that we tried. 
%Please 
Refer to \Cref{sec:4} for more details. 
%Thus:
So:
\small
\begin{equation}
\label{eq:3}
\Pro\Big(\Big(-\log(\widehat{y_{i}}^{(c(\bm{x_{i}}))})\Big)^{1/\beta} < z\Big) = 1 - \exp(-z/\mu)
\end{equation}
\normalsize
The relation between the parameter $\mu$ and the cross-entropy loss value $L$ is given as per \textbf{Lemma 1}.
%\medskip{}

\begin{theorem3*}
\label{lem:1}
Assuming that the expected value of $-\log(\widehat{y}^{(c(\bm{x}))})$ for the model in consideration is approximately equal to the sample average value of this quantity %over the training set 
(which is equal to $L$), we have:
%\[\mu = \Big(\frac{L}{\Gamma(\beta+1)}\Big)^{1/\beta} \text{ where } \Gamma(z) = \int_{0}^{\infty}x^{z-1}e^{-x}dx \text{ (the standard Gamma function)}\]
\[\mu = \Big({L}\Big/{\Gamma(\beta+1)}\Big)^{1/\beta} \text{ where } \Gamma(z) = \int_{0}^{\infty}x^{z-1}e^{-x}dx \text{ (the standard Gamma function)}\]
\end{theorem3*}
\textbf{Proof}:
Let us denote the random variable $(-\log(\widehat{y}^{(c(\bm{x}))}))^{1/\beta}$ by $\widehat{Y}$. Then, $\widehat{Y} \sim \exp(\mu)$ and $\E_{}[\widehat{Y}^{\beta}] = L$. But:
\[\E_{}[\widehat{Y}^{\beta}] = \int_{0}^{\infty} t^{\beta} ({1}/{\mu}) e^{-({t}/{\mu})} dt = \mu^{\beta}\Gamma(\beta+1) = L\]
From this, we get the required result.

%In the next section, we derive the probability of the squared intra-class distance being less than the squared inter-class distance in the $n$-dimensional feature/representation (i.e. the $\phi$) space (due to the network $\mathcal{N}_{c}$) as a function of $L$ to illustrate the separability of the classes. 
In the next section, we derive the probability of the squared inter-class distance being more than the squared intra-class distance times a certain factor $\geq 1$, in the $n$-dimensional feature/representation (i.e. the $\phi$) space (due to %the network 
$\mathcal{N}_{c}$) as a function of $L$, to illustrate the separability of the classes. 

Throughout the rest of this paper, when we say intra/inter-class distance, we mean intra/inter-class distance in the $\phi$ space. Also, $\|\bm{z}\|$ refers to the $\ell_{2}$ norm of the vector $\bm{z}$ and whenever we say {norm}, we mean the {$\ell_{2}$ norm}. Finally, we shall denote the set $\{0,\ldots,(C-1)\}$ by $\bm{[C]}$.

\section{Main Results}
\label{sec:2}
The proofs of all the theorems %(as well as the lemmas and corollaries) 
in this section can be found in the supplementary material.

%We firstly present a theorem which will be used later to obtain an upper bound on the expected value of the squared intra-class distance in the transformed space.
We firstly present a theorem related to the intra-class distance. %in the transformed space.
%\medskip{}

\begin{theorem2*}
Consider a general class, say %$c \in \{0,\ldots,(C-1)\}$
$c \in \bm{[C]}$
, and two randomly chosen points $\bm{x_{1}}$ and $\bm{x_{2}}$ (without loss of generality) belonging to class $c$. Then their transformed representations in the $n$-dimensional $\phi$ space are $\phi(\bm{x_{1}})$ and $\phi(\bm{x_{2}})$, respectively. Let $\Delta \phi(\bm{x}) ^{(c)} = \phi(\bm{x_{1}}) - \phi(\bm{x_{2}})$. Also, let the probabilities of $\bm{x_{1}}$ and $\bm{x_{2}}$ belonging to their ground truth class $c$, predicted by the network be denoted by $\widehat{y_{1}}$ and $\widehat{y_{2}}$, respectively. Finally, consider the $(C-1) \times n$ matrix $\bm{A_{c}}$ whose rows are given by $(\bm{a_{j}} - \bm{a_{c}})$ with $j \in \bm{[C]}-\{c\}$.
Then under the assumption that for each class $j \ne c$,
%the conditional probability of $\bm{x_{1}}$ belonging to $j$ given that it does not belong to $c$ predicted by the network is the same as that for $\bm{x_{2}}$,
${\widehat{y_{1}}^{(j)}}/{(1 - \widehat{y_{1}})} = \widehat{y_{2}}^{(j)}/(1 - \widehat{y_{2}})$ (where $\widehat{y_{i}}^{(j)}$ is the probability of $\bm{x_{i}}$ belonging to class $j$ predicted by the network for $i = \{1,2\}$),
we have:
%\[\E_{}\Big[\|\bm{A_{c}}\Delta \phi(\bm{x}) ^{(c)}\|^{2}\Big] < (C-1) \frac{2\Gamma(2\beta+1)}{(\Gamma(\beta+1))^{2}}L^{2}\]
\small
\[\|\bm{A_{c}}\Delta \phi(\bm{x}) ^{(c)}\|^{2} = (C-1) \log^{2}\Big(\frac{(1/\widehat{y_{1}}) - 1}{(1/\widehat{y_{2}}) - 1}\Big)\]
\normalsize
Also, the complementary cumulative distribution function (ccdf) of $\|\bm{A_{c}}\Delta \phi(\bm{x}) ^{(c)}\|^{2}$ turns out to be:
%\[\Pro_{}(\|\bm{A_{c}}\Delta \phi(\bm{x}) ^{(c)}\|^{2} > \nu (C-1)) = \Bigg\{\int_{0}^{\infty} \Bigg(1 - \exp\Bigg(-\frac{\Big\{\log(1+e^{-\sqrt{\nu}}(e^{(\alpha\mu)^{\beta}}-1))\Big\}^{1/\beta}}{\mu}\Bigg) + \]
%\[\exp\Bigg(-\frac{\Big\{\log(1+e^{\sqrt{\nu}}(e^{(\alpha\mu)^{\beta}}-1))\Big\}^{1/\beta}}{\mu}\Bigg)e^{-\alpha}d\alpha\Bigg\}\]
\small
\[\Pro_{}(\|\bm{A_{c}}\Delta \phi(\bm{x}) ^{(c)}\|^{2} > \nu (C-1)) = 1 -  \int_{0}^{\infty} (e^{-h_{1}(\alpha,-\sqrt{\nu})} - e^{-h_{1}(\alpha,\sqrt{\nu})}) e^{-\alpha}d\alpha  \text{ for } \nu \geq 0,\]
%\[\text{where } h_{1}(w,z) = \frac{\Big\{\log(1+e^{z}(e^{(w\mu)^{\beta}}-1))\Big\}^{1/\beta}}{\mu}\]
\[\text{where } h_{1}(w,z) = \frac{\Big\{\log(1+e^{z}(e^{(w\mu)^{\beta}}-1))\Big\}^{1/\beta}}{\mu} \text{ (and $\mu$ is obtained from \textbf{Lemma 1}).}\]
\normalsize
\end{theorem2*}
%\medskip{}
%The proof of \textbf{Theorem 1} can be found in the supplementary material.
Observe that $\Delta \phi(\bm{x}) ^{(c)}$ is a randomly chosen intra-class distance vector for class $c$. So \textbf{Theorem 1} provides the value as well as the ccdf of the squared norm of the product of the matrix $\bm{A_{c}}$ (which is of course network dependent) and a random intra-class distance vector.
The assumption mentioned in \textbf{Theorem 1} can be %also 
interpreted as follows - given that $\bm{x_{1}}$ and $\bm{x_{2}}$ belong to the same %ground truth 
class $c$, for each class $j \ne c$, the conditional probability of $\bm{x_{1}}$ belonging to class $j$ given that it does not belong to class $c$, predicted by the network, is the same as that for $\bm{x_{2}}$. We acknowledge that this %assumption 
might not be valid for all points (such as adversarial examples) but we %argue
assume that it %should hold 
holds approximately for a large number of points belonging to the same class. %Further, this relatively reasonable assumption makes our ensuing analysis somewhat simpler and without it, the analysis becomes intractable. 
Further, this 
%relatively moderate 
assumption enables us to do some kind of analysis and without it, the problem becomes completely intractable.
Unfortunately, even with this assumption, we could not simplify the ccdf integral further and had to resort to numerical methods to %obtain its value.
evaluate it.
%\medskip{}
%\textbf{Proof}: \textbf{ADD!!}

Next, we present a theorem analogous to \textbf{Theorem 1} for the inter-class distance. %vector in the transformed space.
%\medskip{}

\begin{theorem11*}
Consider two distinct classes, say $c$ and $c' \in \bm{[C]}$. Also, consider two randomly chosen points $\bm{x}$ and $\bm{x'}$ (without loss of generality) belonging to classes $c$ and $c'$, respectively.
Their transformed representations in the $n$-dimensional $\phi$ space are $\phi(\bm{x})$ and $\phi(\bm{x'})$, respectively.
Let $\Delta \phi(\bm{x}) ^{(c,c')} = \phi(\bm{x}) - \phi(\bm{x'})$.
Also, let the probabilities of $\bm{x}$ and $\bm{x'}$ belonging to their respective ground truth classes $c$ and $c'$, predicted by the network be denoted by $\widehat{y}$ and $\widehat{y}'$, respectively. Finally, consider the $(C-1) \times n$ matrix $\bm{A_{c}}$ % (same as in \textbf{Theorem 1} if $c$ is the same as in \textbf{Theorem 1})
 whose rows are given by $(\bm{a_{j}} - \bm{a_{c}})$ with $j \in \bm{[C]}-\{c\}$. Then under the assumption of \textbf{Theorem 1}, for some constant $\kappa_{c,c'} \geq 1$, we have:
\small
\[\|\bm{A_{c}}\Delta \phi(\bm{x}) ^{(c,c')}\|^{2} \geq (C-1) \log^{2}\Bigg(\frac{((\kappa_{c,c'}-1)/{\widehat{y}'}) + 1}{(({1}/{\widehat{y}'}) - 1)(({1}/{\widehat{y}}) - 1)}\Bigg) \]
\normalsize
Also, the ccdf of $\|\bm{A_{c}}\Delta \phi(\bm{x}) ^{(c,c')}\|^{2}$ turns out to be:
\small
\[\Pro_{}\Big(\|\bm{A_{c}}\Delta \phi(\bm{x}) ^{(c,c')}\|^{2} > \nu (C-1)\Big) \geq 1 -  \int_{0}^{\infty} (e^{-h_{2}(\alpha,-\sqrt{\nu})} - e^{-h_{2}(\alpha,\sqrt{\nu})}) e^{-\alpha}d\alpha  \text{ for } \nu \geq 0,\]
%\[\text{where } h_{2}(w,z) = \frac{\Big\{\log(1+e^{z}\Big(\kappa_{c,c'}-1 + \frac{\kappa_{c,c'}}{e^{(w\mu)^{\beta}}-1}\Big))\Big\}^{1/\beta}}{\mu}\]
\[\text{where } h_{2}(w,z) = \frac{\Big\{\log(1+e^{z}\Big(\kappa_{c,c'}-1 + \frac{\kappa_{c,c'}}{e^{(w\mu)^{\beta}}-1}\Big))\Big\}^{1/\beta}}{\mu} \text{ (and $\mu$ is obtained from \textbf{Lemma 1}).}\]
\normalsize
\end{theorem11*}
%\medskip{}
%The proof of \textbf{Theorem 2} can be found in the supplementary material.
%Observe that $\Delta \phi(\bm{x}) ^{(c,c')}$ is a randomly chosen inter-class distance vector between classes $c$ and $c'$. So \textbf{Theorem 2} provides a lower bound on the expected value of the squared norm of the matrix $\bm{A_{c}}$ times the inter-class distance vector.
Observe that $\Delta \phi(\bm{x}) ^{(c,c')}$ is a randomly chosen inter-class distance vector between classes $c$ and $c'$. So \textbf{Theorem 2} provides a lower bound for the value as well as the ccdf of the squared norm of the product of $\bm{A_{c}}$ (network dependent and same as in \textbf{Theorem 1} if $c$ is the same as in \textbf{Theorem 1}) and a random inter-class distance vector. %Here also, we could not simplify the ccdf integral further and had to resort to numerical methods to obtain its value.
%Note that $\kappa_{c,c'}$ turns out to be 
%It turns out that $\kappa_{c,c'}$ is equal to 
$\kappa_{c,c'}$ turns out to be 
the inverse of the conditional probability of $\bm{x'}$ belonging to class $c$ given that it does not belong to class $c'$ %(which is its ground truth class)
(ground truth class of $\bm{x'}$), predicted by the network. %Thus, $\kappa_{c,c'}$ itself is a function of $\alpha$.
Recall from the assumption of \textbf{Theorem 1} (and the discussion below it) that $\kappa_{c,c'}$ is assumed to be a constant for all points within %the same class 
$c'$. 
We further provide a corollary for the special case when all the classes other than the ground truth class are equally similar/dissimilar to each other.
%\medskip{}

\begin{theorem6*}
In \textbf{Theorem 2}, if all classes other than the ground truth class are equally similar/dissimilar to each other, then $\kappa_{c,c'} = (C-1)$ $\forall$ $c \ne c'$. %for all pairs of classes $(c,c')$.
\end{theorem6*}
%\medskip{}
%\textbf{Proof}: \textbf{ADD!!} 
%Here also, we could not simplify the ccdf integral further and had to resort to numerical methods to obtain its value.
Here also, we could not simplify the ccdf integral further and had to resort to numerical methods. %to obtain its value.

\begin{figure}[!h]
\centering 
	\includegraphics[width = 70 mm, height = 52.5 mm]{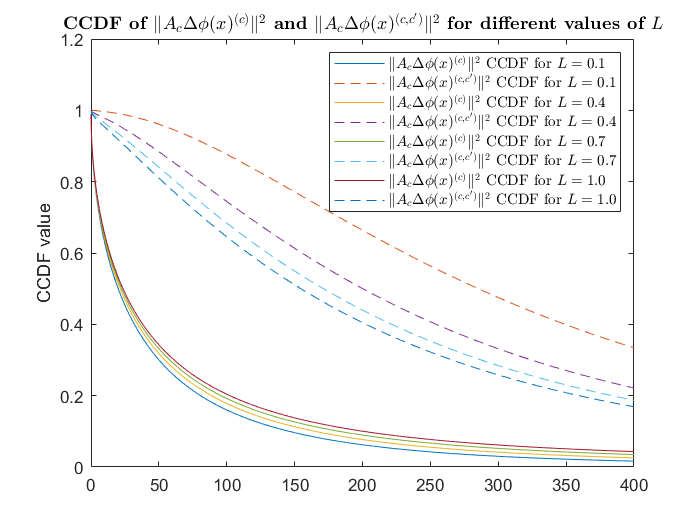}
%\caption{Plots of the ccdf of $\|\bm{A_{c}}\Delta \phi(\bm{x}) ^{(c)}\|^{2}$ and $\|\bm{A_{c}}\Delta \phi(\bm{x}) ^{(c,c')}\|^{2}$ (referred to as \enquote{Intra-Class CCDF...} and \enquote{Inter-Class CCDF...} in the legend, respectively) for $L = 0.1,0.4,0.7$ and $1.0$ with $\beta = 4$, $C = 10$ and $\kappa_{c,c'} = C-1 = 9$.}
\caption{Plots of the ccdf of $\|\bm{A_{c}}\Delta \phi(\bm{x}) ^{(c)}\|^{2}$ and $\|\bm{A_{c}}\Delta \phi(\bm{x}) ^{(c,c')}\|^{2}$ for $L = 0.1,0.4,0.7$ and $1.0$ with $\beta = 4$, $C = 10$ and $\kappa_{c,c'} = C-1 = 9$.}
\label{fig:A0}
\end{figure}

We now show the ccdfs of $\|\bm{A_{c}}\Delta \phi(\bm{x}) ^{(c)}\|^{2}$ and $\|\bm{A_{c}}\Delta \phi(\bm{x}) ^{(c,c')}\|^{2}$ for visual comparison (since it is very difficult to compare them analytically) in \Cref{fig:A0}. Observe that the ccdf of $\|\bm{A_{c}}\Delta \phi(\bm{x}) ^{(c,c')}\|^{2}$ is significantly above that of $\|\bm{A_{c}}\Delta \phi(\bm{x}) ^{(c)}\|^{2}$ which indicates that $\|\bm{A_{c}}\Delta \phi(\bm{x}) ^{(c,c')}\|^{2}$ is more likely to be larger than $\|\bm{A_{c}}\Delta \phi(\bm{x}) ^{(c)}\|^{2}$. Also notice the variation of the ccdfs with respect to $L$. When $L$ increases, the ccdf of $\|\bm{A_{c}}\Delta \phi(\bm{x}) ^{(c)}\|^{2}$ increases while that of $\|\bm{A_{c}}\Delta \phi(\bm{x}) ^{(c,c')}\|^{2}$ decreases. This makes sense intuitively since for higher loss values, we expect the separation of the classes to be worse than that at a lower loss value.

\textbf{Theorem 1} and \textbf{Theorem 2} provide individual results related to the intra-class and inter-class distance vectors. Using these two theorems, we next present a theorem to (probabilistically) compare the values of $\|\bm{A_{c}}\Delta \phi(\bm{x}) ^{(c,c')}\|^{2}$ and $\|\bm{A_{c}}\Delta \phi(\bm{x}) ^{(c)}\|^{2}$.
%\medskip{}

\begin{theorem15*}
Consider a randomly chosen point $\bm{x_{1}}$ belonging to class $c \in \bm{[C]}$. Now consider two points, $\bm{x_{2}}$ also belonging to class $c$ and $\bm{x_{3}}$ belonging to class $c' \ne c$. Their transformed representations in the $n$-dimensional $\phi$ space are $\phi(\bm{x_{1}})$, $\phi(\bm{x_{2}})$ and $\phi(\bm{x_{3}})$, respectively.
Let $\Delta \phi(\bm{x}) ^{(c)} = \phi(\bm{x_{1}}) - \phi(\bm{x_{2}})$ and $\Delta \phi(\bm{x}) ^{(c,c')} = \phi(\bm{x_{1}}) - \phi(\bm{x_{3}})$. %Also recall the matrix $\bm{A_{c}}$ as defined in \textbf{Theorem 1} as well as \textbf{Theorem 2}, $\kappa_{c,c'}$ in \textbf{Theorem 2}, and the assumption in \textbf{Theorem 1}. 
Also 
recall %the matrix 
$\bm{A_{c}}$ as defined in \textbf{Theorem 1} and \textbf{Theorem 2}, the assumption in \textbf{Theorem 1}, $\kappa_{c,c'}$ in \textbf{Theorem 2} and $\mu$ %as obtained 
from \textbf{Lemma 1}. 
Then for any $\gamma > 1$:
\small
\[\Pro_{}\Big(\|\bm{A_{c}}\Delta \phi(\bm{x}) ^{(c,c')}\|^{2} > \gamma\|\bm{A_{c}}\Delta \phi(\bm{x}) ^{(c)}\|^{2}\Big) \geq \]
\[\int_{\alpha_{1} = 0}^{\infty}\int_{\alpha_{2} = 0}^{\infty}|e^{-h_{3}(\alpha_{1},\alpha_{2},-\sqrt{\gamma})} - e^{-h_{3}(\alpha_{1},\alpha_{2},\sqrt{\gamma})}|e^{-\alpha_{2}}e^{-\alpha_{1}}d\alpha_{2}d\alpha_{1} = b_{A}(\gamma,L)\]
\[\text{where } h_{3}(p,q,r) = \frac{1}{\mu}\Bigg\{\log(1+\Big(e^{(p\mu)^{\beta}}-1\Big)^{\frac{1}{1+\frac{1}{r}}}\Big(\kappa_{c,c'} - 1 + \frac{\kappa_{c,c'}}{e^{(q\mu)^{\beta}}-1}\Big)^{\frac{{1}/{r}}{1+\frac{1}{r}}})\Bigg\}^{1/\beta}.\]
\normalsize
%\[\text{and } h_{4}(p,q,r) = \frac{1}{\mu}\Bigg\{\log(1+\Big(e^{(p\mu)^{\beta}}-1\Big)^{\frac{r}{r+1}}\Big(\kappa_{c,c'} - 1 + \frac{\kappa_{c,c'}}{e^{(q\mu)^{\beta}}-1}\Big)^{-\frac{1}{r-1}})\Bigg\}^{1/\beta}.\]
\end{theorem15*}
%\medskip{}

So \textbf{Theorem 3} provides a lower bound, i.e. $b_{A}(\gamma,L)$, on the probability with which the squared norm of the product of the matrix
$\bm{A_{c}}$ and a random inter-class distance vector (between one point in $c$ and another one in $c'$) is more than the squared norm of the product of %the matrix 
$\bm{A_{c}}$ and a random intra-class distance vector (between the aforementioned point in $c$ and another point in $c$ itself), times a certain factor ($\gamma$) $> 1$.
%Once again, the double integral in \textbf{Theorem 3} could not be evaluated analytically due to which we had to evaluate it numerically. 
Once again, the double integral in \textbf{Theorem 3} had to be evaluated numerically. 

\begin{figure}[!h]
\centering 
    \subfloat[Variation vs. $L$]{
	\includegraphics[width = 70 mm, height = 52.5 mm]{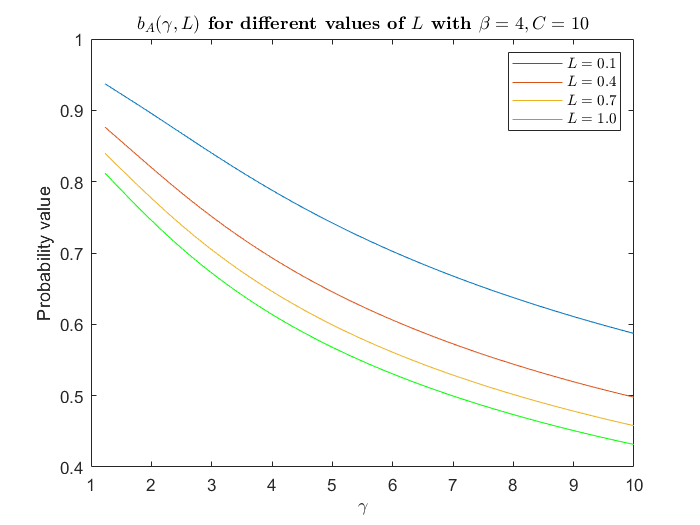}
	\label{fig:A1_a}
	}
	\subfloat[Variation vs. $C$]{
	\includegraphics[width = 70 mm, height = 52.5 mm]{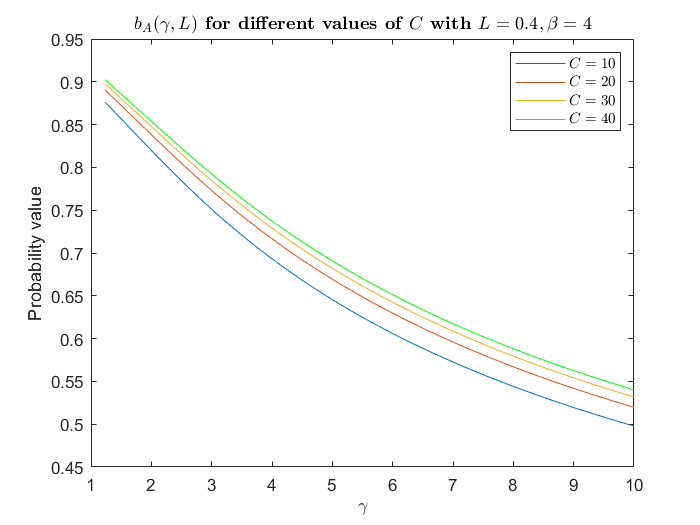}
	\label{fig:A1_b}
	}
\caption{(a) Plots of $b_{A}(\gamma,L)$ for $L = 0.1,0.4,0.7,1.0$ with $\beta = 4$, $C = 10$ and $\kappa_{c,c'} = C-1 = 9$.\\
(b) Plots of $b_{A}(\gamma,L)$ for $C = 10,20,30,40$ with $\beta = 4$, $L = 0.4$ and $\kappa_{c,c'} = C-1$.}
\label{fig:A1}
\end{figure}

In \Cref{fig:A1_a}, we show the variation of $b_{A}(\gamma,L)$ vs. $L$ for fixed values of $\beta = 4$, $C = 10$ and $\kappa_{c,c'} = C-1$. Observe that $b_{A}(\gamma,L)$ decreases as $L$ increases which is consistent with our intuition that separability of the classes is more pronounced for smaller loss values. Similarly, in \Cref{fig:A1_b}, we show the variation of $b_{A}(\gamma,L)$ vs. $C$ for fixed values of $\beta = 4$ and $L = 0.4$. We used $\kappa_{c,c'} = C-1$ for all the four values of $C$. In this case, we observe that $b_{A}(\gamma,L)$ increases as $C$ increases, which is expected since we used $\kappa_{c,c'} = C-1$ (which increases as $C$ increases).

%Now we shall use \textbf{Theorem 3} to derive a lower bound for the
However, we would like to estimate the 
probability with which the squared norm of the inter-class distance vector is more than the squared norm of the intra-class distance vector times a certain factor $\geq 1$.
%However, we would like to have bounds on the expected value of the squared intra-class and inter-class distances itself. 
To do this, we shall use \textbf{Theorem 3} and make %an assumption on the distribution of the entries of $\bm{A_{c}}$ which is as follows:
the following assumption:
%This assumption enables us to bound the squared intra-class and inter-class distances in a small envelope around the suitably re-scaled (by the same constant) value of the squared norm of the product of $\bm{A_{c}}$ and the intra-class and inter-class distance vectors respectively. %The assumption 
%It is as follows:
%\medskip{}
\begin{theorem12*}
The entries of the matrix $\bm{A_{c}}$ are zero mean i.i.d Gaussian random variables.
\end{theorem12*}
\vspace{-2 mm}
This assumption 
%\textbf{Assumption 1} 
%This 
enables us to bound the squared intra-class and inter-class distances in a small envelope around the suitably re-scaled (by the same constant) value of the squared norm of the product of $\bm{A_{c}}$ and the intra-class and inter-class distance vectors respectively.
%Although this assumption may not be perfectly valid,
Although not perfectly valid,
similar assumptions have been used in other works (\cite{gabrie2018entropy,pennington2017geometry}) too.
With the above assumption in mind, we state our main theorem:
%With \textbf{Assumption 1} in mind, we state our main theorem:
%We now state our main theorem:
%\medskip{}
\begin{theorem19*}
Let $F_{C}^{\chi}(u)$ be the cdf of a chi-squared random variable with $(C-1)$ degrees of freedom evaluated at $(C-1)u$. Then, under \textbf{Assumption 1}, the same settings as in \textbf{Theorem 3} and with the function $b_{A}$ as defined in \textbf{Theorem 3}, we have for any $\gamma \geq 1$:
\small
\[\Pro_{}\Big(\|\Delta \phi(\bm{x}) ^{(c,c')}\|^{2} > \gamma \|\Delta\phi(\bm{x}) ^{(c)}\|^{2}\Big) \geq %\int_{0}^{1}\int_{0}^{1} b_{A}\Big(\frac{1+\epsilon_{1}}{1-\epsilon_{2}}\Big)f_{\chi}(1+\epsilon_{1})f_{\chi}(1-\epsilon_{2}) d \epsilon_{1} d \epsilon_{2} = b_{c}(L)
\max_{0 < \epsilon_{1},\epsilon_{2} < 1} b_{A}\Big(\gamma \frac{1+\epsilon_{1}}{1-\epsilon_{2}},L\Big)(F_{{C}}^{\chi}(1+\epsilon_{1}) - F_{C}^{\chi}(1-\epsilon_{2})) = b(\gamma, L)\]
%(F^{\chi}(1+\epsilon_{1},C) - F^{\chi}(1-\epsilon_{2},C)) = b(\gamma, L)\]
\[\geq \max_{0 < \epsilon_{1},\epsilon_{2} < 1} 
%\max_{\epsilon_{1},\epsilon_{2} \in (0,1)} 
b_{A}\Big(\gamma \frac{1+\epsilon_{1}}{1-\epsilon_{2}},L\Big) \Big(1 - %\exp\Big(-\epsilon_{1}^{2}\frac{(C-1)}{4}\Big)
\exp\Big(-(1+\epsilon_{1}-\sqrt{1+2\epsilon_{1}})\frac{(C-1)}{2}\Big)
- \exp\Big(-\epsilon_{2}^{2}\frac{(C-1)}{4}\Big)\Big)\]
\normalsize
\end{theorem19*}
Therefore \textbf{Theorem 4} provides a lower bound, i.e. $b(\gamma,L)$, on the probability with which (the square of) the inter-class distance is more than (the square of) the intra-class distance, times a certain factor, $\gamma \geq 1$. %Also notice that $f_{\chi}(u)$ is equal to the pdf of a chi-squared random variable with $(C-1)$ degrees of freedom at $u(C-1)$.
The bound on the last line of \textbf{Theorem 4} is obtained using certain concentration inequalities for chi-squared random variables.
Throughout the rest of this paper (except in the proof of \textbf{Theorem 4}), we shall consider the case of $\gamma = 1$. For this, let $b_{c}(L) \triangleq{} b(1,L)$. Thus, $b_{c}(L)$ is a lower bound on the probability with which the inter-class distance is more than the intra-class distance.

\begin{figure}[!h]
\centering 
	\includegraphics[width = 70 mm, height = 52.5 mm]{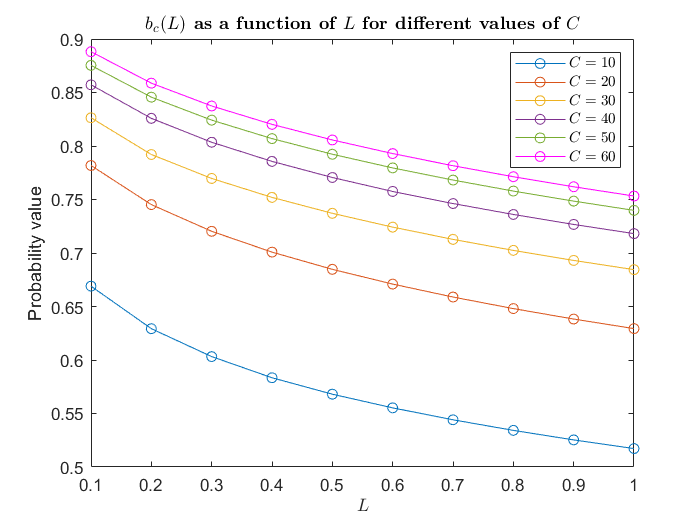}
\caption{Approximate values of $b_{c}(L)$ vs. $L$ for $C = 10,20,30,40,50,60$ with $\kappa_{c,c'} = C-1$ and $\beta = 4$. %fixed to $4$.
We discretized $\epsilon_{1},\epsilon_{2} \in (0,1)$ in steps of $0.1$ each to report the approximate values of $b_{c}(L)$.
}
\label{fig:A2}
\end{figure}

\Cref{fig:A2} shows the variation of $b_{c}(L)$ as a function of $L$ for six different values of $C$. %\textbf{COMMENTS!!!}
In \Cref{fig:A2}, observe that the value of $b_{c}(L)$ decreases as $L$ increases which is what we expect since a higher higher loss value implies poorer separation of the classes or in other words the inter-class distance is less likely to be more than the intra-class distance.
Also, as the number of classes increases, the value of $b_{c}(L)$ also increases which makes sense since $b_{A}(\gamma,L)$ increases as $C$ increases (see \Cref{fig:A1_b}) and %obviously $b_{1}(\epsilon)$, $b_{2}(\epsilon)$ both increase as $C$ increases (recall that $k = C-1$ in \textbf{Lemma 2}).
so does $F_{{C}}^{\chi}(1+\epsilon_{1}) - F_{C}^{\chi}(1-\epsilon_{2})$.
%\medskip{}

{We provide %MATLAB 
code to compute $b(\gamma,L)$ (and $b_A(\gamma,L)$) in the supplementary material.}

%This concludes the section of our theoretical results.
We next provide a theorem on the expected per-class accuracy of the network model in consideration, as a function of $L$, to relate model accuracy with the separability of classes.

\begin{theorem20*}
Let there be $N_{c}$ examples belonging to class $c \in \bm{[C]}$. Under the assumption of \textbf{Theorem 1}, define $\kappa_{c}^{*} \triangleq{} \min_{c' \ne c}\kappa_{c',c}$ ($\kappa_{c',c}$'s are the same as in \textbf{Theorem 2}). 
Then the expected number of examples in class $c$ correctly classified, say $N_{c}^{\text{corr}}$, is: %(the lower bound holds regardless of the validity of the assumption):
\small
\[N_{c}^{\text{corr}} = N_{c}\Big(1 - \exp\Big(-\Big(\frac{\Gamma(\beta+1)\log(1+\kappa_{c}^{*})}{L}\Big)^{1/\beta}\Big)\Big) \geq N_{c}\Big(1 - \exp\Big(-\Big(\frac{\Gamma(\beta+1)\log(2)}{L}\Big)^{1/\beta}\Big)\Big).\]
\normalsize
The lower bound provided above holds regardless of the validity of the assumption in \textbf{Theorem 1}.
\end{theorem20*}
\vspace{-1mm}
%\medskip{}
%In \textbf{Theorem 5}, observe 
Observe that the expected per-class accuracy $(N_{c}^{\text{corr}}/N_{c})$ is a decreasing function of $L$ (as expected) 
%Also, higher the value of $\kappa_{c}^{*}$ for a class $c$, better is its per-class accuracy.
and an increasing function of $\kappa_{c}^{*}$ (i.e. classes with higher $\kappa_{c}^{*}$ will have better per-class accuracy).

%Therefore
So we conclude that a lower loss value leads to better separation of the classes (from \Cref{fig:A2} and \textbf{Theorem 4}) as well as better accuracy (from \textbf{Theorem 5}). %which is consistent with our intuition.

\textbf{Dependence of the obtained results on the dataset and network architecture}: All the theorems stated before depend on the properties of the dataset as well as the network. Firstly, all the results are a function of the cross-entropy loss value which depends on the choice of the network %(for instance, a very simple architecture may not be able to fit the data very well, thereby leading to a higher loss value than that obtained using a more complicated architecture) 
(a very simple architecture may not be able to fit the data very well leading to a higher loss value than that obtained using a more complicated architecture) 
%as compared to a more complicated architecture)
as well as on the dataset (%for instance, 
the same network may perform well on one dataset but not too well on another dataset). Secondly, the value of $\beta$ also plays a critical role in all the obtained expressions and it might depend on the dataset as well as the network architecture (refer to \Cref{fig:1} and the discussion above it in \Cref{sec:4}). In this paper, we do not investigate the variation of $\beta$ or how it affects the obtained expressions. %in depth though.
%in too much detail.
Finally, there is dependence on the number of classes (%which 
can be seen in \Cref{fig:A2}) %as well as in \Cref{fig:A1_b}) 
and also on the value of $\kappa_{c,c'}$ (%which we have 
simply assumed to be $C-1$ in our plots) %- one could think of them as properties of the dataset itself.
%which one could think of as being properties of the dataset itself.
%both of which could be thought of as being properties of the dataset itself.
even though both of them could be thought of as being properties of the dataset itself.
\section{Experiments}
\label{sec:4}
%\subsection{Distribution of }
Due to length constraints on the manuscript, we are able to describe the results of our experiments on only two datasets ({CIFAR-10} and {MNIST}) here. We also performed experiments on two synthetically generated datasets (named {SYN-1} and {SYN-2}) which can be found in the supplementary material. %\textbf{SUPPLEMENTARY MATERIAL}.

%Firstly, we mention some related relevant details for CIFAR-10 and MNIST as well as the (neural network) model that was fitted on it.
Firstly, we mention some relevant details for the CIFAR-10 and MNIST experiments.
%We next show the plots of the distribution of $\Big(-\log(\widehat{y_{i}}^{(c(\bm{x_{i}}))})\Big)^{1/\beta}$ ($\beta$ being specific for each dataset and the corresponding model) along with their closest fit exponential distribution. 
%After that, we present tables containing the probability with which the inter-class distance is more than the intra-class distance for the two datasets.
%\medskip{}

\textbf{CIFAR-10}: %In this case, as explained earlier, we tried with two values of $n$ which were 512 and 1024 (the rest of the network was kept the same, only the final layer dimension was changed). 
%The first dataset that we considered is the well known CIFAR-10 dataset. 
We fitted a deep convolutional neural network (the entire architecture can be found in the supplementary material) having a fully connected final layer with Softmax activation and $n = 512$. %The loss value on the test set after training for \textbf{X!!} epochs was $L = 0.4516$. 
We trained the model for 50 epochs and the corresponding loss value over the test set was $L = 0.4516$. 
%For our network, $n$ (i.e. the final layer dimension) was 512. 
%For this case, we observed that $\beta = 4$ results in %the closest resembling exponential distribution. 
%a closely resembling exponential distribution. 
For this case, we observed that with $\beta = 4$, the distribution of $\Big(-\log(\widehat{y_{i}}^{(c(\bm{x_{i}}))})\Big)^{1/\beta}$ closely resembles an exponential distribution.

\begin{figure}[!h]
\centering 
\subfloat[\textbf{CIFAR-10 with $\beta = 4$}]{
    \label{fig:1_a}
	\includegraphics[width=57.6mm, height=43.2mm]{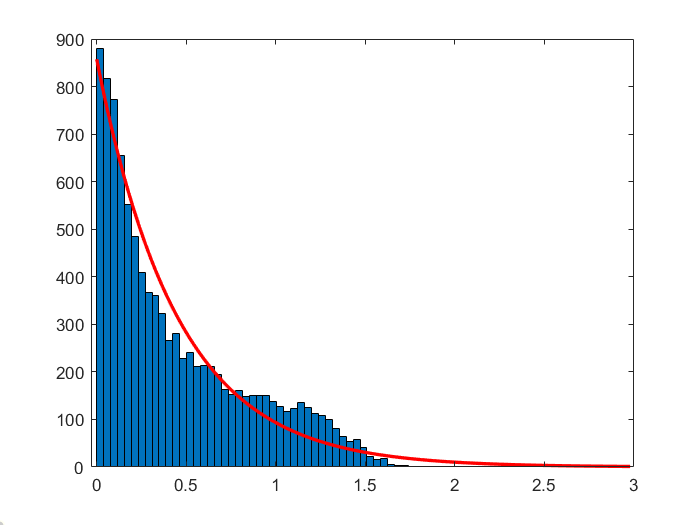}}
\subfloat[\textbf{MNIST with $\beta = 4$}]{
    \label{fig:1_b}
	\includegraphics[width=57.6mm, height=43.2mm]{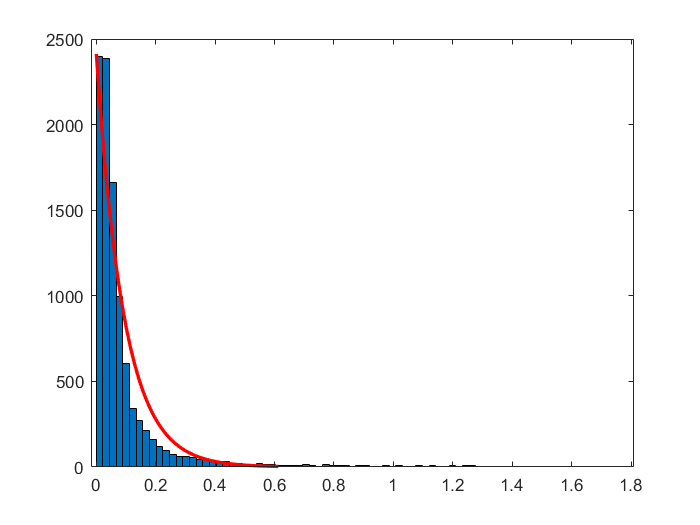}} 
\\ 
\subfloat[\textbf{SYN-1 with $\beta = 2$}]{
    \label{fig:1_c}
	\includegraphics[width=57.6mm, height=43.2mm]{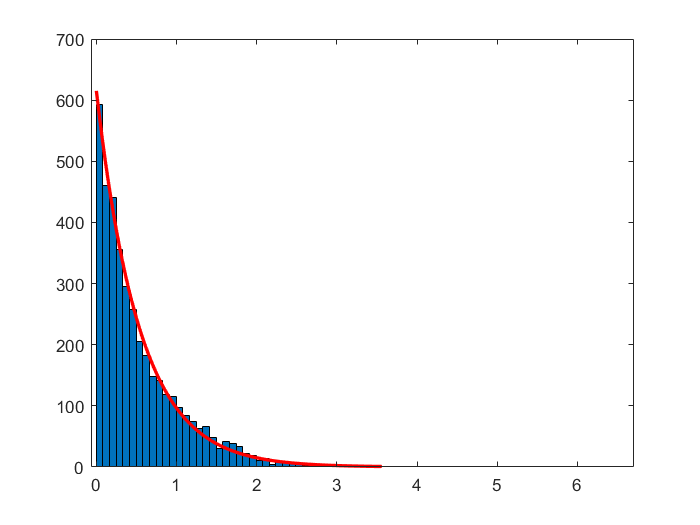}}
\subfloat[\textbf{SYN-2 with %$\beta = 1.18$
$\beta = 1.4$}]{
    \label{fig:1_d}
	\includegraphics[width=57.6mm, height=43.2mm]%{log_y_test_synthetic_third.png}
	{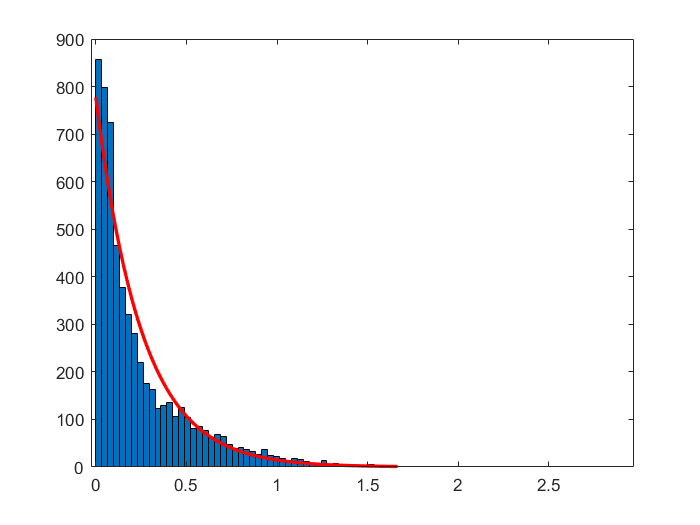}
	}
%\caption{Histogram of $\Big(-\log(\widehat{y_{i}}^{(c(\bm{x_{i}}))})\Big)^{1/\beta}$ along with the closest fit exponential distribution for CIFAR-10 ($n=512$) and MNIST ($n=128$) with $\beta = 4$, SYN-1 ($n = 80$) with $\beta = 2$ and SYN-2 ($n = 80$) with $\beta = 1.18$, respectively, over the test set.}
%\caption{Histogram of $\Big(-\log(\widehat{y_{i}}^{(c(\bm{x_{i}}))})\Big)^{1/\beta}$ along with the closest fit exponential distribution for CIFAR-10 ($n=512$) and MNIST ($n=128$) with $\beta = 4$ respectively, over the test set.}
\caption{Histograms of $\Big(-\log(\widehat{y_{i}}^{(c(\bm{x_{i}}))})\Big)^{1/\beta}$ along with the closest fit exponential distribution for the 4 datasets, over the test set.}
\label{fig:1}
\end{figure}

\textbf{MNIST}: %Next, we experimented with the famous MNIST dataset.
In this case, we fitted a shallow convolutional neural network (the entire architecture and some more details are in the supplementary material) having a fully connected final layer with Softmax activation and $n = 128$. The test set loss value after training the model for 12 epochs was $L = 0.0298$. %in this case.
%The value of $n$ used was $128$.
Just as in the previous case, we observed that even here, $\beta = 4$ results in a closely resembling exponential distribution. 
\Cref{fig:1_a} and \Cref{fig:1_b} contain the plots of the distribution of $\Big(-\log(\widehat{y_{i}}^{(c(\bm{x_{i}}))})\Big)^{1/\beta}$ along with the closest fit exponential distribution for CIFAR-10 and MNIST with $\beta = 4$, over the test set (size of which was 10000 for both datasets). 
The value of $\beta$ probably depends on the dataset, the network architecture and perhaps even the number of classes. For instance, in the case of {SYN-1} and {SYN-2} which are similar datasets with 20 classes each and had shallow fully connected neural networks fitted onto them (more details about these datasets can found in the supplementary material as mentioned earlier), the optimal values of $\beta$ turned out to be $2$ and %$1.18$
$1.4$, respectively.
In \Cref{fig:1_c} and \Cref{fig:1_d}, we show the distribution of $\Big(-\log(\widehat{y_{i}}^{(c(\bm{x_{i}}))})\Big)^{1/\beta}$ along with the closest fit exponential distribution for SYN-1 with $\beta = 2$ and SYN-2 with 
$\beta = 1.4$, respectively, over the test set (size of which was 4000 and 6000 for SYN-1 and SYN-2, respectively). 
%\textbf{The plots of the distribution over the training set can be found in the supplementary material??}
%supplementary material.}

\begin{table}[!h]
\scalebox{0.9}{
\centering
   \subfloat[MNIST \label{tab_1_MNIST}]{
     %\tiny
     \centering
     \begin{tabular}{|l|l|l|l|}
        %\large
        \hline
        %\rule{0pt}{10pt}
                       Class 1 ($c_{1}$) & Class 2 ($c_{2}$) & $p_{1}$ & $p_{2}$     \\
                       \hline
                       %\rule{0pt}{10pt}
        %0 & 1 & 0.9841 & 0.9996 \\
        0 & 1 & 0.9902 & 0.9998 \\
        \hline
        %1 & 2 & 0.9997 & 0.9232 \\
        1 & 2 & 0.9990 & 0.9145 \\
        \hline
        %4 & 9 & 0.9708 & 0.9703 \\
        4 & 9 & 0.9660 & 0.9687 \\
        \hline
        %8 & 0 & 0.9777 & 0.9946 \\
        8 & 0 & 0.9816 & 0.9944 \\
        \hline
        %7 & 1 & 0.9412 & 0.9961 \\
        7 & 1 & 0.9529 & 0.9973 \\
        \hline
        %2 & 3 & 0.9708 & 0.9819 \\
        2 & 3 & 0.9836 & 0.9844 \\
        \hline
        %3 & 8 & 0.9865 & 0.9789 \\
        3 & 8 & 0.9890 & 0.9876 \\
        \hline
        %5 & 6 & 0.9805 & 0.9960 \\
        5 & 6 & 0.9735 & 0.9896 \\
        \hline
        %6 & 8 & 0.9856 & 0.9746 \\
        6 & 8 & 0.9835 & 0.9725 \\
        \hline
        %9 & 8 & 0.9905 & 0.9384 \\
        9 & 8 & 0.9875 & 0.9391 \\
        \hline
    \end{tabular}
   }
   \hspace{0.05 in}
   \subfloat[CIFAR-10 \label{tab_1_CIFAR}]{
     %\tiny
     \centering
     \begin{tabular}{|l|l|l|l|}
        %\large
        \hline
        %\rule{0pt}{10pt}
                       Class 1 ($c_{1}$) & Class 2 ($c_{2}$) & $p_{1}$ & $p_{2}$     \\
                       \hline
                       %\rule{0pt}{10pt}
        %0 & 2 & 0.6371 & 0.7772 \\
        0 & 2 & 0.7255 & 0.8176 \\
        \hline
        %1 & 9 & 0.6526 & 0.7974 \\
        1 & 9 & 0.7376 & 0.8430 \\
        \hline
        %3 & 4 & 0.6781 & 0.6722 \\
        3 & 4 & 0.7369 & 0.7408 \\
        \hline
        %4 & 7 & 0.7878 & 0.6623 \\
        4 & 7 & 0.8289 & 0.7595 \\
        \hline
        %2 & 6 & 0.6886 & 0.7192 \\
        2 & 6 & 0.7668 & 0.8577 \\
        \hline
        5 & 6 & 0.8253 & 0.8433 \\
        \hline
        %1 & 4 & 0.8945 & 0.9308 \\
        1 & 4 & 0.8933 & 0.9535 \\
        \hline
        %6 & 3 & 0.7439 & 0.7470 \\
        6 & 3 & 0.7600 & 0.7752 \\
        \hline
        %7 & 5 & 0.6176 & 0.7460 \\
        7 & 5 & 0.6755 & 0.7835 \\
        \hline
        %8 & 1 & 0.4311 & 0.9375 \\
        8 & 1 & 0.5461 & 0.9406 \\
        \hline
        %9 & 5 &  &  \\
        %\hline
    \end{tabular}
   }

}
\caption{Sample probabilities of inter-class distance being more than the intra-class distance for (a) MNIST with $L = 0.0298$ and (b) CIFAR-10 with $L = 0.4516$. $p_{1}$ and $p_{2}$ are as defined in the text.}
\label{tab1_in}
\end{table}
Finally, \Cref{tab_1_MNIST} and \Cref{tab_1_CIFAR} show the probability of the inter-class distance being more than the intra-class distance for MNIST and CIFAR-10 respectively, over the test set. For every dataset, we chose 10 pairs of classes and for each pair of classes (denote the two classes in a pair by $c_{1}$ and $c_{2}$), we considered the transformed representations of 200 random points in each class (denote these by $\{\phi(\bm{x_{1}^{(1)}}),\phi(\bm{x_{1}^{(2)}}),\ldots,\phi(\bm{x_{1}^{(200)}})\}$ and $\{\phi(\bm{x_{2}^{(1)}}),\phi(\bm{x_{2}^{(2)}}),\ldots,\phi(\bm{x_{2}^{(200)}})\}$ for $c_{1}$ and $c_{2}$, respectively). Next, for each $c_{i}$ such that $i = \{1,2\}$ and for each $j$ such that $1 \leq j \leq 100$, we considered 100 intra-class distance vectors (denote these by $\bm{d_{i}^{(j,k)}} = \phi(\bm{x_{i}^{(j)}}) - \phi(\bm{x_{i}^{(k+100)}})$ for $1 \leq k \leq 100$) and 100 inter-class distance vectors (denote these by $\bm{d_{1,2}^{(j,k')}} = \phi(\bm{x_{1}^{(j)}}) - \phi(\bm{x_{2}^{(k')}})$ for $1 \leq k' \leq 100$). Then, $p_{1}$ and $p_{2}$ (see \Cref{tab_1_MNIST} and \Cref{tab_1_CIFAR}) are mathematically defined as follows:
\small
\[p_{i} = \sum_{j = 1}^{100} \sum_{k = 1}^{100} \sum_{k' = 1}^{100} \bm{1} \Big(\|\bm{d_{1,2}^{(j,k')}}\| > \|\bm{d_{i}^{(j,k)}}\|\Big) \Big/ 100^{3} \text{ for } i = \{1,2\}\] 
\[\text{ where } \bm{1}(.) \text{ is the indicator function and evaluates to 1 if its argument is true else it evaluates to 0.}\]
\normalsize
In other words, $p_{1}$ and $p_{2}$ are the sample probabilities of the inter-class distance %(i.e. distance between one point in $c_{1}$ and another one in $c_{2}$) 
between $c_{1}$ and $c_{2}$
being more than the intra-class distance of $c_{1}$ and $c_{2}$, %(i.e. distance between two points belonging to the same class, $c_{1}$ and $c_{2}$), 
respectively.

%\textbf{DISCUSS IT AND COMPARE WITH THEORY!}
In \Cref{tab_1_MNIST}, observe that $p_{1}$ and $p_{2}$ values are very close to $1$ and the corresponding value of $L$ is 0.0298. The value of $b_{c}(0.0298)$ with $C = 10, \kappa_{c,c'} = C-1$ and $\beta = 4$ turns out to be $0.7251$ approximately. 
In \Cref{tab_1_CIFAR}, where $L = 0.4516$,
the values of $p_{1}$ and $p_{2}$ are much lower than $1$ and mostly in the range of $0.7-0.85$. %and the corresponding value of $L$ is 0.4516. 
The value of $b_{c}(0.4516)$ with $C = 10, \kappa_{c,c'} = C-1$ and $\beta = 4$ is $0.5750$ approximately.
Thus, the obtained lower bounds %on the probability values
are consistent with the observed values.
%The values of $b_{c}(0.0298)$ and $b_{c}(0.4516)$ with $C = 10, \kappa_{c,c'} = C-1, \beta = 4$ turn out to be approximately $0.7251$ and $0.5750$, respectively. %\textbf{COMPARE WITH THEORY!} 

Further experiments on SYN-1 and SYN-2 are described in the supplementary material.

\section{Conclusions}
\label{sec:5}
%\textbf{FILL IT!}
In this paper, we have attempted to mathematically quantify the separability of classes with the cross-entropy loss function by deriving a lower bound on the probability with which the inter-class distance is more than the intra-class distance (in the feature space learnt by a neural network) as a function of the loss value. 
%The constant $\beta$ plays an important role in our analysis and in this work, we have not analyzed how it depends on the network structure or the dataset or the number of classes. A future line of work could be to analyze the variation of $\beta$ and its impact on the probability values. 
The constant $\beta$ plays an important role in our results. A future line of work could be to analyze the variation of $\beta$ and its impact on the probability values. 
%Also, we have derived \textbf{Theorem 4} by assuming that the entries of $\bm{A_{c}}$ are zero mean i.i.d Gaussian random variables. One could possibly look to extend the work to the case of the entries of $\bm{A_{c}}$ having a sub-Gaussian distribution by using concentration inequalities for sub-Gaussian random variables, similar to those that have been used to obtain \textbf{Theorem 4}. 
One could also possibly look to relax \textbf{Assumption 1} and extend the work to the case of the entries of $\bm{A_{c}}$ having a sub-Gaussian distribution by using concentration inequalities for sub-Gaussian random variables, similar to those that have been used to obtain \textbf{Theorem 4}. 
Finally, we hope that our techniques can be extended or suitably modified to do the same task for other loss functions too.

\bibliographystyle{plain}%siamplain}
\bibliography{references}

\begin{thebibliography}{10}

\bibitem{chen2018virtual}
Binghui Chen, Weihong Deng, and Haifeng Shen.
\newblock Virtual class enhanced discriminative embedding learning.
\newblock In {\em Advances in Neural Information Processing Systems}, pages
  1942--1952, 2018.

\bibitem{cortes1995support}
Corinna Cortes and Vladimir Vapnik.
\newblock Support-vector networks.
\newblock {\em Machine learning}, 20(3):273--297, 1995.

\bibitem{gabrie2018entropy}
Marylou Gabri{\'e}, Andre Manoel, Cl{\'e}ment Luneau, Nicolas Macris, Florent
  Krzakala, Lenka Zdeborov{\'a}, et~al.
\newblock Entropy and mutual information in models of deep neural networks.
\newblock In {\em Advances in Neural Information Processing Systems}, pages
  1821--1831, 2018.

\bibitem{gao2018margin}
Riqiang Gao, Fuwei Yang, Wenming Yang, and Qingmin Liao.
\newblock Margin loss: Making faces more separable.
\newblock {\em IEEE Signal Processing Letters}, 25(2):308--312, 2018.

\bibitem{he2016identity}
Kaiming He, Xiangyu Zhang, Shaoqing Ren, and Jian Sun.
\newblock Identity mappings in deep residual networks.
\newblock In {\em European conference on computer vision}, pages 630--645.
  Springer, 2016.

\bibitem{JL_Gaussian4}
Sham Kakade and Greg Shakhnarovich.
\newblock {\em Random Projections}, 2009.
\newblock
  \url{https://ttic.uchicago.edu/~gregory/courses/LargeScaleLearning/lectures/jl.pdf}.

\bibitem{krizhevsky2012imagenet}
Alex Krizhevsky, Ilya Sutskever, and Geoffrey~E Hinton.
\newblock Imagenet classification with deep convolutional neural networks.
\newblock In {\em Advances in neural information processing systems}, pages
  1097--1105, 2012.

\bibitem{laurent2000adaptive}
Beatrice Laurent and Pascal Massart.
\newblock Adaptive estimation of a quadratic functional by model selection.
\newblock {\em Annals of Statistics}, pages 1302--1338, 2000.

\bibitem{liu2017sphereface}
Weiyang Liu, Yandong Wen, Zhiding Yu, Ming Li, Bhiksha Raj, and Le~Song.
\newblock Sphereface: Deep hypersphere embedding for face recognition.
\newblock In {\em Proceedings of the IEEE conference on computer vision and
  pattern recognition}, pages 212--220, 2017.

\bibitem{liu2016large}
Weiyang Liu, Yandong Wen, Zhiding Yu, and Meng Yang.
\newblock Large-margin softmax loss for convolutional neural networks.
\newblock In {\em ICML}, volume~2, page~7, 2016.

\bibitem{luo2019mathcal}
Yan Luo, Yongkang Wong, Mohan Kankanhalli, and Qi~Zhao.
\newblock $\mathcal {G} $-softmax: Improving intra-class compactness and
  inter-class separability of features.
\newblock {\em arXiv preprint arXiv:1904.04317}, 2019.

\bibitem{pennington2017geometry}
Jeffrey Pennington and Yasaman Bahri.
\newblock Geometry of neural network loss surfaces via random matrix theory.
\newblock In {\em Proceedings of the 34th International Conference on Machine
  Learning-Volume 70}, pages 2798--2806. JMLR. org, 2017.

\bibitem{simonyan2014very}
Karen Simonyan and Andrew Zisserman.
\newblock Very deep convolutional networks for large-scale image recognition.
\newblock {\em arXiv preprint arXiv:1409.1556}, 2014.

\bibitem{sun2014deep}
Yi~Sun, Yuheng Chen, Xiaogang Wang, and Xiaoou Tang.
\newblock Deep learning face representation by joint
  identification-verification.
\newblock In {\em Advances in neural information processing systems}, pages
  1988--1996, 2014.

\bibitem{szegedy2017inception}
Christian Szegedy, Sergey Ioffe, Vincent Vanhoucke, and Alexander~A Alemi.
\newblock Inception-v4, inception-resnet and the impact of residual connections
  on learning.
\newblock In {\em Thirty-First AAAI Conference on Artificial Intelligence},
  2017.

\bibitem{zhou2019separability}
Liguo Zhou, Zhongyuan Wang, Yimin Luo, and Zixiang Xiong.
\newblock Separability and compactness network for image recognition and
  superresolution.
\newblock {\em IEEE transactions on neural networks and learning systems},
  2019.

\end{thebibliography}
\clearpage

\section{Supplementary Material}
\label{sec:3}

\subsection{Proof of \textbf{Theorem 1}}
\label{sec:3_1}
Firstly, we prove \textbf{Theorem 1}. Before proving it, we restate it for the reader's convenience.
\medskip{}

\begin{theorem2*}
Consider a general class, say $c \in \bm{[C]}$, and two randomly chosen points $\bm{x_{1}}$ and $\bm{x_{2}}$ (without loss of generality) belonging to class $c$. Then their transformed representations in the $n$-dimensional $\phi$ space are $\phi(\bm{x_{1}})$ and $\phi(\bm{x_{2}})$, respectively. Let $\Delta \phi(\bm{x}) ^{(c)} = \phi(\bm{x_{1}}) - \phi(\bm{x_{2}})$. Also, let the probabilities of $\bm{x_{1}}$ and $\bm{x_{2}}$ belonging to their ground truth class $c$, predicted by the network be denoted by $\widehat{y_{1}}$ and $\widehat{y_{2}}$, respectively. Finally, consider the $(C-1) \times n$ matrix $\bm{A_{c}}$ whose rows are given by $(\bm{a_{j}} - \bm{a_{c}})$ with $j \in \bm{[C]}-\{c\}$. Then under the assumption that for each class $j \ne c$,
${\widehat{y_{1}}^{(j)}}/{(1 - \widehat{y_{1}})} = \widehat{y_{2}}^{(j)}/(1 - \widehat{y_{2}})$ (where $\widehat{y_{i}}^{(j)}$ is the probability of $\bm{x_{i}}$ belonging to class $j$ predicted by the network for $i = \{1,2\}$), we have:
%\[\E_{}\Big[\|\bm{A_{c}}\Delta \phi(\bm{x}) ^{(c)}\|^{2}\Big] < (C-1) \frac{2\Gamma(2\beta+1)}{(\Gamma(\beta+1))^{2}}L^{2}\]
\small
\[\|\bm{A_{c}}\Delta \phi(\bm{x}) ^{(c)}\|^{2} = (C-1) \log^{2}\Big(\frac{(1/\widehat{y_{1}}) - 1}{(1/\widehat{y_{2}}) - 1}\Big)\]
\normalsize
Also, the complementary cumulative distribution function (ccdf) of $\|\bm{A_{c}}\Delta \phi(\bm{x}) ^{(c)}\|^{2}$ turns out to be:
\small
%\[\Pro_{}(\|\bm{A_{c}}\Delta \phi(\bm{x}) ^{(c)}\|^{2} > \nu (C-1)) = \Bigg\{\int_{0}^{\infty} \Bigg(1 - \exp\Bigg(-\frac{\Big\{\log(1+e^{-\sqrt{\nu}}(e^{(\alpha\mu)^{\beta}}-1))\Big\}^{1/\beta}}{\mu}\Bigg) + \]
%\[\exp\Bigg(-\frac{\Big\{\log(1+e^{\sqrt{\nu}}(e^{(\alpha\mu)^{\beta}}-1))\Big\}^{1/\beta}}{\mu}\Bigg)e^{-\alpha}d\alpha\Bigg\}\]
\[\Pro_{}(\|\bm{A_{c}}\Delta \phi(\bm{x}) ^{(c)}\|^{2} > \nu (C-1)) = 1 -  \int_{0}^{\infty} (e^{-h_{1}(\alpha,-\sqrt{\nu})} - e^{-h_{1}(\alpha,\sqrt{\nu})}) e^{-\alpha}d\alpha  \text{ for } \nu \geq 0,\]
\[\text{where } h_{1}(w,z) = \frac{\Big\{\log(1+e^{z}(e^{(w\mu)^{\beta}}-1))\Big\}^{1/\beta}}{\mu} \text{ (and $\mu$ is obtained from \textbf{Lemma 1}).}\]
\normalsize
\end{theorem2*}
\medskip{}

\textbf{Proof}: We have-
\[{1}/{\widehat{y_{i}}^{(c)}} = 1 + \sum_{k \ne c} \exp((\bm{a_{k}}-\bm{a_{c}})^{T}\phi(\bm{x_{i}}) + (b_{c}-b_{k})) \text{ for } i = \{1,2\}.\]
Now, ${1}/{\widehat{y_{1}}^{(c)}} = 1/\widehat{y_{1}} \text{ (as mentioned in the theorem) } \Rightarrow \exp((\bm{a_{j}}-\bm{a_{c}})^{T}\phi(\bm{x_{1}}) + (b_{j}-b_{c})) = \eta_{j}^{(1)}({1}/{\widehat{y_{1}}}-1)$ for some $ \eta_{j}^{(1)}$ such that $0 \leq \eta_{j}^{(1)} \leq 1$ and $\sum_{k \ne c} \eta_{k}^{(1)} = 1$. Thus:
\begin{equation}
    \label{eq:4}
    (\bm{a_{j}}-\bm{a_{c}})^{T}\phi(\bm{x_{1}}) + (b_{j}-b_{c}) = \log(\eta_{j}^{(1)}({1}/{\widehat{y_{1}}}-1)).
\end{equation}
%Thus, $(\bm{a_{j}}-\bm{a_{c}})^{T}\phi(\bm{x_{1}}) + (b_{j}-b_{c}) = \log(\eta_{j}^{(1)}({1}/{\widehat{y_{1}}}-1))$.

Similarly, for some $ \eta_{j}^{(2)}$'s such that $0 \leq \eta_{j}^{(2)} \leq 1$ and $\sum_{k \ne c} \eta_{k}^{(2)} = 1$, we have -
\begin{equation}
\label{eq:5}
    (\bm{a_{j}}-\bm{a_{c}})^{T}\phi(\bm{x_{2}}) + (b_{j}-b_{c}) = \log(\eta_{j}^{(2)}({1}/{\widehat{y_{2}}}-1)).
\end{equation}
%$(\bm{a_{j}}-\bm{a_{c}})^{T}\phi(\bm{x_{2}}) + (b_{j}-b_{c}) = \log(\eta_{j}^{(2)}({1}/{\widehat{y_{2}}}-1))$ for some $ \eta_{j}^{(2)}$ such that $0 \leq \eta_{j}^{(2)} \leq 1$ and $\sum_{k \ne c} \eta_{k}^{(2)} = 1$.

Subtracting (\ref{eq:5}) from (\ref{eq:4}) gives us:
%Subtracting the above two equations, we get -
%$(\bm{a_{c}}-\bm{a_{j}})^{T}(\phi(\bm{x_{1}})-\phi(\bm{x_{2}})) = \log(\frac{\eta_{j}^{(1)}(P-1)}{\eta_{j}^{(2)}(Q-1)})$.
%\begin{comment}
\begin{equation}
\label{eq:6}
(\bm{a_{j}}-\bm{a_{c}})^{T}(\phi(\bm{x_{1}})-\phi(\bm{x_{2}})) = \log(\frac{\eta_{j}^{(1)}({1}/{\widehat{y_{1}}}-1)}{\eta_{j}^{(2)}({1}/{\widehat{y_{2}}}-1)}).
\end{equation}
%\end{comment}
%So from (\ref{eq:6}), we have:
Therefore, we have:
\begin{equation}
\label{eq:7}
%\|\bm{A_{c}}\Delta \phi(\bm{x}) ^{(c)}\|^{2} = \sum_{k \ne c}\{(\bm{a_{c}}-\bm{a_{k}})^{T}(\phi(\bm{x_{1}})-\phi(\bm{x_{2}}))\}^{2} = \sum_{k \ne c}\Bigg\{\log(\frac{\eta_{k}^{(1)}(P-1)}{\eta_{k}^{(2)}(Q-1)})\Bigg\}^{2}.
\|\bm{A_{c}}\Delta \phi(\bm{x}) ^{(c)}\|^{2} = \sum_{k \ne c}\{(\bm{a_{k}}-\bm{a_{c}})^{T}(\phi(\bm{x_{1}})-\phi(\bm{x_{2}}))\}^{2} = \sum_{k \ne c}\log^{2}\Bigg(\frac{\eta_{k}^{(1)}({1}/{\widehat{y_{1}}}-1)}{\eta_{k}^{(2)}({1}/{\widehat{y_{2}}}-1)}\Bigg).
\end{equation}
Now, since $\bm{x_{1}}$ and $\bm{x_{2}}$ belong to the same class $c$, we assume that $\eta_{j}^{(1)} = \eta_{j}^{(2)}$ for all $j \ne c$.
%Now we assume that $\eta_{j}^{(1)} = \eta_{j}^{(2)}$ for all $j \ne c$ (since $\bm{x_{1}}$ and $\bm{x_{2}}$ belong to the same class $c$).

From (\ref{eq:4}), we have:
\[\eta_{j}^{(1)} =  \frac{\exp(\bm{a_{j}}^{T}\phi(\bm{x_{1}}) + b_{j})}{\exp(\bm{a_{c}}^{T}\phi(\bm{x_{1}}) + b_{c})({1}/{\widehat{y_{1}}}-1)} = \frac{\widehat{y_{1}}^{(j)}}{\widehat{y_{1}}({1}/{\widehat{y_{1}}}-1)} = \frac{\widehat{y_{1}}^{(j)}}{1 - \widehat{y_{1}}}\]
From the above equation, it can be also seen that $\eta_{j}^{(1)}$ is the conditional probability of of $\bm{x_{1}}$ belonging to class $j$ given that it does not belong to class $c$, predicted by the network.

Similarly, from (\ref{eq:5}), we have:
\[\eta_{j}^{(2)} =  \frac{\widehat{y_{2}}^{(j)}}{1 - \widehat{y_{2}}}\]
Thus, $\eta_{j}^{(1)} = \eta_{j}^{(2)}$ for all $j \ne c \Rightarrow {\widehat{y_{1}}^{(j)}}/{(1 - \widehat{y_{1}})} = {\widehat{y_{2}}^{(j)}}/{(1 - \widehat{y_{2}})}$ for all $j \ne c$. It also implies that the conditional probability of $\bm{x_{1}}$ belonging to class $j$ given that it does not belong to class $c$, predicted by the network, is the same as that for $\bm{x_{2}}$, for all $j \ne c$.
We acknowledge that this assumption might not be valid for all points (such as adversarial examples) but we %argue that it should 
assume that it holds approximately for a large number of points belonging to the same class. %Further, this assumption makes our ensuing analysis somewhat simpler and without it, the analysis becomes intractable. 
Further, without this assumption, our ensuing analysis becomes intractable. 
%In other words, we assume that the extent to which every class (other than the ground truth class) seems likely/unlikely to $\bm{x_{1}}$ and $\bm{x_{2}}$ (which is quantified by the $\eta_{j}^{(1)}$'s and the  $\eta_{j}^{(2)}$'s) is roughly the same for both of them. This is the assumption that has been mentioned in the theorem statement.

%We acknowledge the fact that this assumption might not be valid for all examples (such as adversarial examples) but we argue that it should hold (approximately) for a large number of examples. Further, this relatively reasonable assumption makes our ensuing analysis somewhat simpler and without it, the analysis becomes intractable. 

Therefore, under the aforementioned assumption, we get:
\begin{equation}
\label{eq:8}
\|\bm{A_{c}}\Delta \phi(\bm{x}) ^{(c)}\|^{2} = (C-1)\log^{2}\Bigg(\frac{({1}/{\widehat{y_{1}}}-1)}{({1}/{\widehat{y_{2}}}-1)}\Bigg).
\end{equation}
This proves the first part of \textbf{Theorem 1}.

%To prove the second part of \textbf{Theorem 1}, recall that $\Big(\log(1/\widehat{y}^{(c(\bm{x}))})\Big)^{1/\beta}$ approximately follows an exponential distribution with mean $\mu$ ($\beta$ is a constant $> 1$). Keeping this in mind, let us introduce a useful reparameterization as follows:
%\[\Big(\log(1/\widehat{y}^{(c(\bm{x}))})\Big)^{1/\beta} = \alpha \mu \text{ where } \alpha \sim \exp(1) \Rightarrow 1/\widehat{y}^{(c(\bm{x}))} = e^{{\alpha\mu}^{\beta}} \text{ where } \alpha \sim \exp(1)\]
Next using (\ref{eq:8}), we have: $\Pro_{}(\|\bm{A_{c}}\Delta \phi(\bm{x}) ^{(c)}\|^{2} < \nu (C-1)) =$
\begin{equation}
\label{eq:9}
    %\Pro_{}(\|\bm{A_{c}}\Delta \phi(\bm{x}) ^{(c)}\|^{2} < \nu (C-1)) =
    \int_{1}^{\infty} \Pro_{}\Big(1+e^{-\sqrt{\nu}}\Big(\frac{1}{\widehat{y_{2}}}-1\Big)<\frac{1}{\widehat{y_{1}}}<1+e^{\sqrt{\nu}}\Big(\frac{1}{\widehat{y_{2}}}-1\Big)\Big)  %e^{-\alpha}d\alpha
    f\Big(\frac{1}{\widehat{y_{2}}}\Big)d\Big(\frac{1}{\widehat{y_{2}}}\Big)
    \text{ where } f(.) \text{ is the pdf of } \Big(\frac{1}{\widehat{y_{2}}}\Big).
\end{equation}
Now recall that $\Big(\log(1/\widehat{y}^{(c(\bm{x}))})\Big)^{1/\beta}$ approximately follows an exponential distribution with mean $\mu$ whose value (as a function of $L$) is obtained from \textbf{Lemma 1} in the main paper. Keeping this in mind, let us introduce a useful reparameterization as follows:
\[\Big(\log(1/\widehat{y}^{(c(\bm{x}))})\Big)^{1/\beta} = \alpha \mu \text{ where } \alpha \sim \exp(1) \Rightarrow 1/\widehat{y}^{(c(\bm{x}))} = e^{{(\alpha\mu)}^{\beta}} \text{ where } \alpha \sim \exp(1)\]

%Then with the above reparameterization, (\ref{eq:9}), reduces to:
Then with the above reparameterization, let ${1}/{\widehat{y_{1}}} = e^{(\alpha_{1}\mu)^{\beta}}$ and ${1}/{\widehat{y_{2}}} = e^{(\alpha_{2}\mu)^{\beta}}$ where $\alpha_{1},\alpha_{2}$ are i.i.d  $\exp(1)$. So we have:
\begin{multline}
    \label{eq:10}
    \Pro_{}\Big(1+e^{-\sqrt{\nu}}\Big(\frac{1}{\widehat{y_{2}}}-1\Big)<\frac{1}{\widehat{y_{1}}}<1+e^{\sqrt{\nu}}\Big(\frac{1}{\widehat{y_{2}}}-1\Big)\Big) = 
    \Pro_{}(h_{1}(\alpha_{2},-\sqrt{\nu})<\alpha_{1}<h_{1}(\alpha_{2},\sqrt{\nu})) \\
    =
    (e^{-h_{1}(\alpha_{2},-\sqrt{\nu})} - e^{-h_{1}(\alpha_{2},\sqrt{\nu})})
    \text{ where } h_{1}(w,z) = \frac{\Big\{\log(1+e^{z}(e^{(w\mu)^{\beta}}-1))\Big\}^{1/\beta}}{\mu}
\end{multline}
%\[\text{ where } h_{1}(w,z) = \frac{\Big\{\log(1+e^{z}(e^{(w\mu)^{\beta}}-1))\Big\}^{1/\beta}}{\mu}\]
We also have:
\begin{equation}
    \label{eq:11}
    f\Big(\frac{1}{\widehat{y_{2}}}\Big)d\Big(\frac{1}{\widehat{y_{2}}}\Big) = e^{-\alpha_{2}}d\alpha_{2}
\end{equation}
Now using (\ref{eq:10}) and (\ref{eq:11}) in (\ref{eq:9}), we finally get:
\begin{equation}
\label{eq:12}
    \Pro_{}(\|\bm{A_{c}}\Delta \phi(\bm{x}) ^{(c)}\|^{2} < \nu (C-1)) = 
    \int_{0}^{\infty} (e^{-h_{1}(\alpha_{2},-\sqrt{\nu})} - e^{-h_{1}(\alpha_{2},\sqrt{\nu})}) e^{-\alpha_{2}}d\alpha_{2}
\end{equation}
%\end{equation}
Lastly, replacing $\alpha_{2}$ by $\alpha$ in (\ref{eq:12}), we obtain the second part of \textbf{Theorem 1}:
\begin{multline}
\label{eq:13}
    \Pro_{}(\|\bm{A_{c}}\Delta \phi(\bm{x}) ^{(c)}\|^{2} > \nu (C-1)) = 1 -
    \int_{0}^{\infty} (e^{-h_{1}(\alpha,-\sqrt{\nu})} - e^{-h_{1}(\alpha,\sqrt{\nu})}) e^{-\alpha}d\alpha
    \\
    \text{with $h_{1}()$ as defined in (\ref{eq:10}).}
%\end{equation}
\end{multline}
%\[\text{with $h_{1}()$ as defined in \ref{eq:10}.}\]

This finishes the proof of \textbf{Theorem 1}. Unfortunately, we could not simplify the integral in (\ref{eq:13}) further.

\subsection{Proof of \textbf{Theorem 2}}
\label{sec:3_2}
Next, we restate \textbf{Theorem 2} and then prove it. %followed by its proof.

\begin{theorem11*}
Consider two distinct classes, say $c$ and $c' \in \bm{[C]}$. Also, consider two randomly chosen points $\bm{x}$ and $\bm{x'}$ (without loss of generality) belonging to classes $c$ and $c'$, respectively.
Their transformed representations in the $n$-dimensional $\phi$ space are $\phi(\bm{x})$ and $\phi(\bm{x'})$, respectively.
Let $\Delta \phi(\bm{x}) ^{(c,c')} = \phi(\bm{x}) - \phi(\bm{x'})$.
Also, let the probabilities of $\bm{x}$ and $\bm{x'}$ belonging to their respective ground truth classes, $c$ and $c'$, predicted by the network be denoted by $\widehat{y}$ and $\widehat{y}'$, respectively. Finally, consider the $(C-1) \times n$ matrix $\bm{A_{c}}$ % (same as in \textbf{Theorem 1})
 whose rows are given by $(\bm{a_{j}} - \bm{a_{c}})$ with $j \in \bm{[C]}-\{c\}$. %Then for some $\kappa_{c,c'} \geq 1$, we have:
Then under the assumption of \textbf{Theorem 1}, for some constant $\kappa_{c,c'} \geq 1$, we have:
\small
\[\|\bm{A_{c}}\Delta \phi(\bm{x}) ^{(c,c')}\|^{2} \geq (C-1) \log^{2}\Bigg(\frac{((\kappa_{c,c'}-1)/{\widehat{y}'}) + 1}{(({1}/{\widehat{y}'}) - 1)(({1}/{\widehat{y}}) - 1)}\Bigg) \]
\normalsize
Also, the ccdf of $\|\bm{A_{c}}\Delta \phi(\bm{x}) ^{(c,c')}\|^{2}$ turns out to be:
\small
\[\Pro_{}\Big(\|\bm{A_{c}}\Delta \phi(\bm{x}) ^{(c,c')}\|^{2} > \nu (C-1)\Big) \geq 1 -  \int_{0}^{\infty} (e^{-h_{2}(\alpha,-\sqrt{\nu})} - e^{-h_{2}(\alpha,\sqrt{\nu})}) e^{-\alpha}d\alpha  \text{ for } \nu \geq 0,\]
\[
\text{where } h_{2}(w,z) = \frac{\Big\{\log(1+e^{z}\Big(\kappa_{c,c'}-1 + \frac{\kappa_{c,c'}}{e^{(w\mu)^{\beta}}-1}\Big))\Big\}^{1/\beta}}{\mu} \text{ (and $\mu$ is obtained from \textbf{Lemma 1}).}
\]
\normalsize
\end{theorem11*}
\medskip{}

\textbf{Proof}: Similar to the derivation of (\ref{eq:4}) and (\ref{eq:5}) in the proof of \textbf{Theorem 1}, we have for some $\eta_{j}$'s such that $0 \leq \eta_{j} \leq 1$ and $\sum_{k \ne c} \eta_{k} = 1$:
\begin{equation}
    \label{eq:14}
    (\bm{a_{j}}-\bm{a_{c}})^{T}\phi(\bm{x}) + (b_{j}-b_{c}) = \log(\eta_{j}({1}/{\widehat{y}}-1)).
\end{equation}

Now since $\bm{x'}$ belongs to class $c'$ and the probability of it belonging to $c'$ itself, predicted by the network is assumed to be $\widehat{y}'$ (as stated in the theorem), let us say that the probability of $\bm{x'}$ belonging to class $c$ predicted by the network is ${(1-\widehat{y}')}/{\kappa_{c,c'}}$ where $\kappa_{c,c'} \geq 1$. 
%The constant $\kappa_{c,c'}$ is a measure of the similarity of the two classes $c$ and $c'$. So if $c$ and $c'$ are similar(dissimilar) to each other, then the value of $\kappa_{c,c'}$ is low(high). Additionally, if all the classes are equally similar/dissimilar to each other, then obviously $\kappa_{c,c'} = C-1$. This is exactly the statement of \textbf{Corollary 1}.
It is easy to see that $1/\kappa_{c,c'}$ is equal to the conditional probability of $\bm{x'}$ belonging to class $c$ given that it does not belong to class $c'$ (which is its ground truth class), predicted by the network.
Also recall from the assumption of \textbf{Theorem 1} that $\kappa_{c,c'}$ is assumed to be a constant for all points belonging to $c'$. 
Additionally, if all the classes other than $c'$ are equally similar/dissimilar to each other, then obviously $\kappa_{c,c'} = (C-1)$ for all $c \ne c'$ (since $\sum_{j \ne c'}(1/\kappa_{j,c'}) = 1$ and if all the classes other than $c'$ are equally similar/dissimilar to each other, then that would imply $\kappa_{j,c'}^{-1} = (C-1)^{-1}$ $\forall$ $j \ne c'$). This is exactly the statement of \textbf{Corollary 1}.

So once again, similar to the derivation of (\ref{eq:4}) and (\ref{eq:5}) in the proof of \textbf{Theorem 1}, we have for some $\eta_{j}'$ such that $0 \leq \eta_{j}' \leq 1$ and $\sum_{k \ne c} \eta_{k}' = 1$:
\begin{equation}
    \label{eq:15}
    (\bm{a_{j}}-\bm{a_{c}})^{T}\phi(\bm{x'}) + (b_{j}-b_{c}) = \log(\eta_{j}'\Big(\frac{\kappa_{c,c'}}{(1-\widehat{y}')}-1\Big)) = \log(\eta_{j}'\frac{(\kappa_{c,c'}-1)(1/\widehat{y}')+1}{(1/\widehat{y}')-1}).
\end{equation}

Now subtracting (\ref{eq:15}) from (\ref{eq:14}), we get:
\begin{equation}
    \label{eq:16}
    (\bm{a_{j}}-\bm{a_{c}})^{T}(\phi(\bm{x}) - \phi(\bm{x'})) = \log(\frac{\eta_{j}}{\eta_{j}'}\frac{(({1}/{\widehat{y}'}) - 1)(({1}/{\widehat{y}}) - 1)}{((\kappa_{c,c'}-1)/{\widehat{y}'}) + 1})
\end{equation}

Therefore, we have:
\begin{equation}
\label{eq:17}
\|\bm{A_{c}}\Delta \phi(\bm{x}) ^{(c,c')}\|^{2} = \sum_{k \ne c}\{(\bm{a_{k}}-\bm{a_{c}})^{T}(\phi(\bm{x})-\phi(\bm{x'}))\}^{2} = \sum_{k \ne c}\log^{2}\Big(\frac{\eta_{k}}{\eta_{k}'}\frac{(({1}/{\widehat{y}'}) - 1)(({1}/{\widehat{y}}) - 1)}{((\kappa_{c,c'}-1)/{\widehat{y}'}) + 1}\Big)
\end{equation}

Note that here we do not assume $\eta_{j} = \eta_{j}'$ for $j \ne c$ since $\bm{x}$ and $\bm{x'}$ belong to different classes (whereas the assumption of \textbf{Theorem 1} is for points belonging to the same class).
%Next, 
Instead, for this case, we state a lemma which allows us to provide a %more 
succinct bound. 
\begin{theorem4*}
\label{lem:2}
Consider the function $g(u_{1},\ldots,u_{p},v_{1},\ldots,v_{p}) = \sum_{i=1}^{p} \log^{2}\Big(\frac{u_{i}}{v_{i}}R\Big)$ where $R > 0$ is a constant. Then under the constraints $0 < u_{i},v_{i} < 1$ $\forall i \in \{1,\ldots,p\}$ and $\sum_{i=1}^{p} u_{i} = \sum_{i=1}^{p} v_{i} = 1$, the minimum value of $g$ occurs when $u_{i} = v_{i} = (\frac{1}{p})$ $\forall i 
%\in \{1,\ldots,p\}
$ and the minimum value is equal to $p\log^{2}(R)$.
\end{theorem4*}
%\medskip{}

We defer the proof of \textbf{Lemma 2} for now but it can be found immediately after the end of the proof of \textbf{Theorem 2} in this subsection itself.

Observe that the RHS of (\ref{eq:17}) has the same form as the function $g$ in \textbf{Lemma 2} with $p = C-1$, $R = \frac{(({1}/{\widehat{y}'}) - 1)(({1}/{\widehat{y}}) - 1)}{((\kappa_{c,c'}-1)/{\widehat{y}'}) + 1}$ and $u_{i}$ and $v_{i}$ playing the roles of $\eta_{k}$ and $\eta_{k}'$, respectively. 
Thus using \textbf{Lemma 2}, we have:
\begin{multline}
\label{eq:18}
\|\bm{A_{c}}\Delta \phi(\bm{x}) ^{(c,c')}\|^{2} \geq (C-1)\log^{2}\Big(\frac{(({1}/{\widehat{y}'}) - 1)(({1}/{\widehat{y}}) - 1)}{((\kappa_{c,c'}-1)/{\widehat{y}'}) + 1}\Big) = \\ (C-1) \log^{2}\Big(\frac{((\kappa_{c,c'}-1)/{\widehat{y}'}) + 1}{(({1}/{\widehat{y}'}) - 1)(({1}/{\widehat{y}}) - 1)}\Big)
\end{multline}

This completes the proof of the first part of \textbf{Theorem 2}.

Next using (\ref{eq:18}), we have: $\Pro_{}(\|\bm{A_{c}}\Delta \phi(\bm{x}) ^{(c,c')}\|^{2} < \nu (C-1)) \leq$
\begin{multline}
\label{eq:19}
    \int_{1}^{\infty} \Bigg\{\Pro_{}\Big(1+e^{-\sqrt{\nu}}\Big(\frac{(\kappa_{c,c'}-1)(1/\widehat{y}')+1}{(1/\widehat{y}')-1}\Big)<\frac{1}{\widehat{y}}<1+e^{\sqrt{\nu}}\Big(\frac{(\kappa_{c,c'}-1)(1/\widehat{y}')+1}{(1/\widehat{y}')-1}\Big)\Big)  %e^{-\alpha}d\alpha
    \\
    f\Big(\frac{1}{\widehat{y}'}\Big)d\Big(\frac{1}{\widehat{y}'}\Big)\Bigg\}
    \text{ where } f(.) \text{ is the pdf of } \Big(\frac{1}{\widehat{y}'}\Big).
\end{multline}

Using the reparameterization mentioned in the proof of \textbf{Theorem 1}, let us say that ${1}/{\widehat{y}} = e^{(\alpha\mu)^{\beta}}$ and ${1}/{\widehat{y}'} = e^{(\alpha'\mu)^{\beta}}$ where $\alpha,\alpha'$ are i.i.d  $\exp(1)$. Then we have:
\begin{multline}
    \label{eq:20}
    \Pro_{}\Big(1+e^{-\sqrt{\nu}}\Big(\frac{(\kappa_{c,c'}-1)(1/\widehat{y}')+1}{(1/\widehat{y}')-1}\Big)<\frac{1}{\widehat{y}}<1+e^{\sqrt{\nu}}\Big(\frac{(\kappa_{c,c'}-1)(1/\widehat{y}')+1}{(1/\widehat{y}')-1}\Big)\Big) \\
    =
    \Pro_{}(h_{2}(\alpha',-\sqrt{\nu})<\alpha<h_{2}(\alpha',\sqrt{\nu}))
    =
    (e^{-h_{2}(\alpha',-\sqrt{\nu})} - e^{-h_{2}(\alpha',\sqrt{\nu})}) \\
    \text{ where } h_{2}(w,z) = \frac{\Big\{\log(1+e^{z}\Big(\kappa_{c,c'}-1 + \frac{\kappa_{c,c'}}{e^{(w\mu)^{\beta}}-1}\Big))\Big\}^{1/\beta}}{\mu}
\end{multline}

We also have:
\begin{equation}
    \label{eq:21}
    f\Big(\frac{1}{\widehat{y}'}\Big)d\Big(\frac{1}{\widehat{y}'}\Big) = e^{-\alpha'}d\alpha'
\end{equation}
Now using (\ref{eq:20}) and (\ref{eq:21}) in (\ref{eq:19}), we finally get:
\begin{equation}
\label{eq:22}
    \Pro_{}(\|\bm{A_{c}}\Delta \phi(\bm{x}) ^{(c,c')}\|^{2} < \nu (C-1)) \leq 
    \int_{0}^{\infty} (e^{-h_{2}(\alpha',-\sqrt{\nu})} - e^{-h_{2}(\alpha',\sqrt{\nu})}) e^{-\alpha'}d\alpha'
\end{equation}
%\end{equation}
Lastly, replacing $\alpha'$ by $\alpha$ in (\ref{eq:22}), we obtain the second part of \textbf{Theorem 2}:
%\begin{equation}
\begin{multline}
\label{eq:23}
    \Pro_{}(\|\bm{A_{c}}\Delta \phi(\bm{x}) ^{(c,c')}\|^{2} > \nu (C-1)) \geq 1 - \int_{0}^{\infty} (e^{-h_{2}(\alpha,-\sqrt{\nu})} - e^{-h_{2}(\alpha,\sqrt{\nu})}) e^{-\alpha}d\alpha
    \\
    \text{with $h_{2}()$ as defined in (\ref{eq:20}).}
\end{multline}
%\end{equation}
%\[\text{with $h_{2}(.,.)$ as defined in \ref{eq:20}.}\]

This finishes the proof of the second part of \textbf{Theorem 2}. Unfortunately, we could not simplify the integral in (\ref{eq:23}) further.

Now we prove \textbf{Lemma 2}. We wish to minimize $g(u_{1},\ldots,u_{p},v_{1},\ldots,v_{p}) = \sum_{i=1}^{p} \log^{2}\Big(\frac{u_{i}}{v_{i}}R\Big)$ under the constraints $0 < u_{i},v_{i} < 1$ $\forall i \in \{1,\ldots,p\}$ and $\sum_{i=1}^{p} u_{i} = \sum_{i=1}^{p} v_{i} = 1$. We solve this using the method of Lagrangian multipliers (only imposing the equality constraints). Ideally, we should %use the KKT con
also impose the inequality constraints (i.e. $0 < u_{i},v_{i} < 1$) and use the KKT conditions. However, it turns out that the solution obtained using the method of Lagrangian multipliers satisfies the inequality constraints due to which we need not worry.
Thus, consider: 
\[h(u_{1},\ldots,u_{p},v_{1},\ldots,v_{p},\lambda_{u},\lambda_{v}) = \sum_{i=1}^{p} \log^{2}\Big(\frac{u_{i}}{v_{i}}R\Big) - \lambda_{u}(\sum_{i=1}^{p} u_{i} - 1) + \lambda_{v}(\sum_{i=1}^{p} v_{i} - 1)\]
We have:
\begin{equation}
    \label{eq:20_n}
    \frac{\partial h}{\partial u_{i}} = 2\log(\frac{u_{i}R}{v_{i}})\frac{1}{u_{i}} - \lambda_{u} = 0 \Rightarrow \log(\frac{u_{i}R}{v_{i}}) = \frac{\lambda_{u}u_{i}}{2}
\end{equation}
%\[\frac{\partial h}{\partial u_{i}} = 2\log(\frac{u_{i}R}{v_{i}})\frac{1}{u_{i}} - \lambda_{u} = 0 \Rightarrow \log(\frac{u_{i}R}{v_{i}}) = \frac{\lambda_{u}u_{i}}{2}\]
\begin{equation}
    \label{eq:21_n}
    \frac{\partial h}{\partial v_{i}} = 2\log(\frac{u_{i}R}{v_{i}})\frac{-1}{v_{i}} + \lambda_{v} = 0 \Rightarrow \log(\frac{u_{i}R}{v_{i}}) = \frac{\lambda_{v}v_{i}}{2}
\end{equation}
%\[\frac{\partial h}{\partial v_{i}} = 2\log(\frac{u_{i}R}{v_{i}})\frac{-1}{v_{i}} + \lambda_{v} = 0 \Rightarrow \log(\frac{u_{i}R}{v_{i}}) = \frac{\lambda_{v}v_{i}}{2}\]
From (\ref{eq:20_n}) and (\ref{eq:21_n}), we get:
\begin{equation}
    \label{eq:22_n}
    \frac{u_{i}}{v_{i}} = \frac{\lambda_{v}}{\lambda_{u}} = \alpha \text{ }\forall \text{ }i
\end{equation}
Now using (\ref{eq:22_n}) along with (\ref{eq:20_n}) and (\ref{eq:21_n}), we get:
\begin{equation}
    \label{eq:23_n}
    u_{i} = \frac{2}{\lambda_{u}}\log(\alpha R) \text{ , }
    v_{i} = \frac{2}{\lambda_{v}}\log(\alpha R) \text{ }\forall \text{ }i
\end{equation}
Finally using (\ref{eq:23_n}) along with the constraint that $\sum_{i=1}^{p} u_{i} = 1$ and $\sum_{i=1}^{p} v_{i} = 1$, we get - 
\begin{equation}
    \label{eq:24_n}
    u_{i} = v_{i} = \frac{1}{p} \text{ }\forall \text{ }i
\end{equation}
Thus, the minimum value is obtained when $u_{i} = v_{i} = {1}/{p}$ $\forall \text{ }i$. Observe that the obtained solution automatically satisfies the inequality constraints (i.e. $0 < u_{i},v_{i} < 1$), as mentioned before.
The minimum value obtained for $u_{i} = v_{i} = {1}/{p}$ $\forall \text{ }i$ is obviously equal to $p\log^{2}(R)$. 
This finishes the proof of \textbf{Lemma 2}.

\subsection{Proof of \textbf{Theorem 3}}
\label{sec:3_3}
Now, we restate \textbf{Theorem 3} followed by its proof.
%Now we prove \textbf{Theorem 3}. Before proving it, we restate it for the reader's convenience.
\medskip{}

\begin{theorem15*}
Consider a randomly chosen point $\bm{x_{1}}$ belonging to class $c \in \bm{[C]}$. Now consider two points, $\bm{x_{2}}$ also belonging to class $c$ and $\bm{x_{3}}$ belonging to class $c' \ne c$. Their transformed representations in the $n$-dimensional $\phi$ space are $\phi(\bm{x_{1}})$, $\phi(\bm{x_{2}})$ and $\phi(\bm{x_{3}})$, respectively.
Let $\Delta \phi(\bm{x}) ^{(c)} = \phi(\bm{x_{1}}) - \phi(\bm{x_{2}})$ and $\Delta \phi(\bm{x}) ^{(c,c')} = \phi(\bm{x_{1}}) - \phi(\bm{x_{3}})$. 
\begin{comment}
Also %consider
recall the %$(C-1) \times n$ 
matrix $\bm{A_{c}}$ as defined in \textbf{Theorem 1} as well as \textbf{Theorem 2}, % and 
%the constant 
%$\kappa_{c,c'}$ in \textbf{Theorem 2}, 
%and 
the assumption in \textbf{Theorem 1}, $\kappa_{c,c'}$ in \textbf{Theorem 2} and $\mu$ as obtained from \textbf{Lemma 1}.
\end{comment}
Also recall $\bm{A_{c}}$ as defined in \textbf{Theorem 1} and \textbf{Theorem 2}, the assumption in \textbf{Theorem 1}, $\kappa_{c,c'}$ in \textbf{Theorem 2} and $\mu$ %as obtained 
from \textbf{Lemma 1}. 
Then for any $\gamma > 1$:
\small
\[\Pro_{}\Big(\|\bm{A_{c}}\Delta \phi(\bm{x}) ^{(c,c')}\|^{2} > \gamma\|\bm{A_{c}}\Delta \phi(\bm{x}) ^{(c)}\|^{2}\Big) \geq \]
\[\int_{\alpha_{1} = 0}^{\infty}\int_{\alpha_{2} = 0}^{\infty}|e^{-h_{3}(\alpha_{1},\alpha_{2},-\sqrt{\gamma})} - e^{-h_{3}(\alpha_{1},\alpha_{2},\sqrt{\gamma})}|e^{-\alpha_{2}}e^{-\alpha_{1}}d\alpha_{2}d\alpha_{1} = b_{A}(\gamma,L)\]
\[\text{where } h_{3}(p,q,r) = \frac{1}{\mu}\Bigg\{\log(1+\Big(e^{(p\mu)^{\beta}}-1\Big)^{\frac{1}{1+\frac{1}{r}}}\Big(\kappa_{c,c'} - 1 + \frac{\kappa_{c,c'}}{e^{(q\mu)^{\beta}}-1}\Big)^{\frac{{1}/{r}}{1+\frac{1}{r}}})\Bigg\}^{1/\beta}\]
\normalsize
\end{theorem15*}
\medskip{}

\textbf{Proof}: 
From \textbf{Theorem 1}, we have:
\[\|\bm{A_{c}}\Delta \phi(\bm{x}) ^{(c)}\|^{2} = (C-1) \log^{2}\Big(\frac{(1/\widehat{y_{1}}) - 1}{(1/\widehat{y_{2}}) - 1}\Big) = q_{1}(\widehat{y_{1}},\widehat{y_{2}})\]
\[\text{ where $\widehat{y_{1}}$ and $\widehat{y_{2}}$ are the probabilities predicted by the network of $\bm{x_{1}}$ and $\bm{x_{2}}$ belonging to class $c$.}\]
Also, from \textbf{Theorem 2}, we have:
\[\|\bm{A_{c}}\Delta \phi(\bm{x}) ^{(c,c')}\|^{2} \geq (C-1) \log^{2}\Bigg(\frac{((\kappa_{c,c'}-1)/{\widehat{y_{3}}}) + 1}{(({1}/{\widehat{y_{3}}}) - 1)(({1}/{\widehat{y_{1}}}) - 1)}\Bigg) = q_{2}(\widehat{y_{1}},\widehat{y_{3}})\]
\[\text{ where $\widehat{y_{3}}$ is the probability predicted by the network of $\bm{x_{3}}$ belonging to class $c'$.}\]

Now since $q_{2}(\widehat{y_{1}},\widehat{y_{3}})$ is a lower bound for $\|\bm{A_{c}}\Delta \phi(\bm{x}) ^{(c,c')}\|^{2}$:
\begin{equation}
    \label{eq:24}
    \Pro_{}\Big(\|\bm{A_{c}}\Delta \phi(\bm{x}) ^{(c,c')}\|^{2} > \gamma\|\bm{A_{c}}\Delta \phi(\bm{x}) ^{(c)}\|^{2}\Big) \geq \Pro_{}\Big(q_{2}(\widehat{y_{1}},\widehat{y_{3}}) > \gamma q_{1}(\widehat{y_{1}},\widehat{y_{2}})\Big)
\end{equation}
%\[\Pro_{}\Big(\|\bm{A_{c}}\Delta \phi(\bm{x}) ^{(c,c')}\|^{2} > \gamma\|\bm{A_{c}}\Delta \phi(\bm{x}) ^{(c)}\|^{2}\Big) \geq \Pro_{}\Big(q_{2}(\widehat{y_{1}},\widehat{y_{3}}) > \gamma q_{1}(\widehat{y_{1}},\widehat{y_{2}})\Big)\]
Let us now use the reparameterization introduced in the proof of
\textbf{Theorem 1}. Say, ${1}/{\widehat{y_{1}}} = e^{(\alpha_{1}\mu)^{\beta}}$, ${1}/{\widehat{y_{2}}} = e^{(\alpha_{2}\mu)^{\beta}}$ and ${1}/{\widehat{y_{3}}} = e^{(\alpha_{3}\mu)^{\beta}}$ where $\alpha_{1},\alpha_{2},\alpha_{3}$ are i.i.d  $\exp(1)$. 
We then have -
\begin{equation}
    \label{eq:25}
    q_{1}(\widehat{y_{1}},\widehat{y_{2}}) = \log^{2}\Big(\frac{e^{(\alpha_{1}\mu)^{\beta}} - 1}{e^{(\alpha_{2}\mu)^{\beta}} - 1}\Big) \text{ and } q_{2}(\widehat{y_{1}},\widehat{y_{3}}) = \log^{2}\Bigg(\frac{(\kappa_{c,c'}-1) + \frac{\kappa_{c,c'}}{e^{(\alpha_{3}\mu)^{\beta}} - 1}}{e^{(\alpha_{1}\mu)^{\beta}}-1}\Bigg)
\end{equation}
Using (\ref{eq:25}) and imposing the condition: $q_{2}(\widehat{y_{1}},\widehat{y_{3}}) - \gamma q_{1}(\widehat{y_{1}},\widehat{y_{2}}) > 0$, we get:
%\begin{comment}
\begin{multline}
    \label{eq:26}
      \alpha_{1} \in (\min(h_{3}(\alpha_{2},\alpha_{3},\sqrt{\gamma}),h_{3}(\alpha_{2},\alpha_{3},-\sqrt{\gamma})),\max(h_{3}(\alpha_{2},\alpha_{3},\sqrt{\gamma}),h_{3}(\alpha_{2},\alpha_{3},-\sqrt{\gamma})))
      \\
      \text{where } h_{3}(p,q,r) = \frac{1}{\mu}\Bigg\{\log(1+\Big(e^{(p\mu)^{\beta}}-1\Big)^{\frac{1}{1+\frac{1}{r}}}\Big(\kappa_{c,c'} - 1 + \frac{\kappa_{c,c'}}{e^{(q\mu)^{\beta}}-1}\Big)^{\frac{{1}/{r}}{1+\frac{1}{r}}})\Bigg\}^{1/\beta}
\end{multline}
%\[\text{where } h_{3}(p,q,r) = \frac{1}{\mu}\Bigg\{\log(1+\Big(e^{(p\mu)^{\beta}}-1\Big)^{\frac{1}{1+\frac{1}{r}}}\Big(\kappa_{c,c'} - 1 + \frac{\kappa_{c,c'}}{e^{(q\mu)^{\beta}}-1}\Big)^{\frac{{1}/{r}}{1+\frac{1}{r}}})\Bigg\}^{1/\beta}\]
%\end{comment}
%The above equation is obtained
Imposing the condition: $q_{2}(\widehat{y_{1}},\widehat{y_{3}}) - \gamma q_{1}(\widehat{y_{1}},\widehat{y_{2}}) > 0$, gives us a quadratic inequality in $\log(e^{(\alpha_{1}\mu)^{\beta}} - 1)$ and the coefficients of the inequality are a function of $\log(e^{(\alpha_{2}\mu)^{\beta}} - 1)$ as well as $\log((\kappa_{c,c'}-1) + \frac{\kappa_{c,c'}}{e^{(\alpha_{3}\mu)^{\beta}} - 1})$. Solving the quadratic inequality gives us (\ref{eq:26}). 

Next, we have:
\begin{multline}
    \label{eq:27}
      \Pro_{}(\alpha_{1} \in (\min(h_{3}(\alpha_{2},\alpha_{3},\sqrt{\gamma}),h_{3}(\alpha_{2},\alpha_{3},-\sqrt{\gamma})),\max(h_{3}(\alpha_{2},\alpha_{3},\sqrt{\gamma}),h_{3}(\alpha_{2},\alpha_{3},-\sqrt{\gamma}))))
      \\
      = |\exp({-h_{3}(\alpha_{2},\alpha_{3},-\sqrt{\gamma})}) - \exp({-h_{3}(\alpha_{2},\alpha_{3},\sqrt{\gamma})})|.
\end{multline}

Hence using (\ref{eq:27}), we get:
\begin{equation}
    \label{eq:28}
    \Pro_{}\Big(q_{2}(\widehat{y_{1}},\widehat{y_{3}}) > \gamma q_{1}(\widehat{y_{1}},\widehat{y_{2}})\Big) = \int_{\alpha_{2} = 0}^{\infty}\int_{\alpha_{3} = 0}^{\infty}|e^{-h_{3}(\alpha_{2},\alpha_{3},-\sqrt{\gamma})} - e^{-h_{3}(\alpha_{2},\alpha_{3},\sqrt{\gamma})}|e^{-\alpha_{3}}e^{-\alpha_{2}}d\alpha_{3}d\alpha_{2}
\end{equation}

Finally, using (\ref{eq:24}) along with (\ref{eq:28}) and replacing the variables $\alpha_{3}$ and $\alpha_{2}$ in (\ref{eq:28}) by $\alpha_{2}$ and $\alpha_{1}$ respectively, we get:
\begin{multline}
    \label{eq:29}
    \Pro_{}\Big(\|\bm{A_{c}}\Delta \phi(\bm{x}) ^{(c,c')}\|^{2} > \gamma\|\bm{A_{c}}\Delta \phi(\bm{x}) ^{(c)}\|^{2}\Big) \geq 
    \\
    \int_{\alpha_{1} = 0}^{\infty}\int_{\alpha_{2} = 0}^{\infty}|e^{-h_{3}(\alpha_{1},\alpha_{2},-\sqrt{\gamma})} - e^{-h_{3}(\alpha_{1},\alpha_{2},\sqrt{\gamma})}|e^{-\alpha_{2}}e^{-\alpha_{1}}d\alpha_{2}d\alpha_{1} \text{ with $h_{3}()$ as defined in (\ref{eq:26}).}
\end{multline}

This concludes the proof of \textbf{Theorem 3}. Unfortunately, we could not simplify the double integral in (\ref{eq:29}) any further.

\subsection{Proof of \textbf{Theorem 4}}
\label{sec:3_4}
We now restate and then prove %the last theorem, i.e.
\textbf{Theorem 4}. 
\medskip{}

\begin{theorem19*}
Let $F_{C}^{\chi}(u)$ be the cdf of a chi-squared random variable with $(C-1)$ degrees of freedom evaluated at $(C-1)u$. Then, under \textbf{Assumption 1}, the same settings as in \textbf{Theorem 3} and with the function $b_{A}$ as defined in \textbf{Theorem 3}, we have for any $\gamma \geq 1$:
\small
\[\Pro_{}\Big(\|\Delta \phi(\bm{x}) ^{(c,c')}\|^{2} > \|\Delta\phi(\bm{x}) ^{(c)}\|^{2}\Big) \geq %\int_{0}^{1}\int_{0}^{1} b_{A}\Big(\frac{1+\epsilon_{1}}{1-\epsilon_{2}}\Big)f_{\chi}(1+\epsilon_{1})f_{\chi}(1-\epsilon_{2}) d \epsilon_{1} d \epsilon_{2} = b_{c}(L)
\max_{0 < \epsilon_{1},\epsilon_{2} < 1} b_{A}\Big(\gamma \frac{1+\epsilon_{1}}{1-\epsilon_{2}},L\Big)(F_{{C}}^{\chi}(1+\epsilon_{1}) - F_{C}^{\chi}(1-\epsilon_{2})) = b(\gamma,L)\]
\[\geq \max_{0 < \epsilon_{1},\epsilon_{2} < 1} b_{A}\Big(\gamma \frac{1+\epsilon_{1}}{1-\epsilon_{2}},L\Big) \Big(1 - \exp\Big(-(1+\epsilon_{1}-\sqrt{1+2\epsilon_{1}})\frac{(C-1)}{2}\Big)
- \exp\Big(-\epsilon_{2}^{2}\frac{(C-1)}{4}\Big)\Big)\]
\normalsize
\end{theorem19*}
\medskip{}

\textbf{Proof}:
Firstly, recall our assumption (\textbf{Assumption 1}) that the elements of $\bm{A_{c}}$ are zero mean i.i.d Gaussian random variables. Let $\bm{\widehat{A}_{c}} \triangleq \frac{\bm{A_{c}}}{\sigma_{c}\sqrt{C-1}}$ where $\sigma_{c}^{2}$ is the variance of the elements of ${\bm{A_{c}}}$. Thus, the elements of $\bm{\widehat{A}_{c}} \sim \mathcal{N}(0,\frac{1}{C-1})$.

Now as shown in \cite{JL_Gaussian4}, both $\|\bm{\widehat{A}_{c}} \Delta \phi(\bm{x}) ^{(c)}\|^{2}/\|\Delta \phi(\bm{x}) ^{(c)}\|^{2}$ and $\|\bm{\widehat{A}_{c}} \Delta \phi(\bm{x}) ^{(c,c')}\|^{2}/\|\Delta \phi(\bm{x}) ^{(c,c')}\|^{2}$ are both chi-squared random variables with $(C-1)$ degrees of freedom each (here $\bm{\widehat{A}_{c}}$ is treated as random variable whereas $\Delta \phi(\bm{x}) ^{(c)}$ and $\Delta \phi(\bm{x}) ^{(c,c')}$ are treated as deterministic/non-random quantities). But unfortunately, the two are not independent (we are talking about independence with respect to $\bm{\widehat{A}_{c}}$) of each other. But still, we have the following inequality:
\begin{multline}
    \label{eq:30}
    \Pro_{}\Bigg(\gamma \|\Delta \phi(\bm{x}) ^{(c)}\|^{2} <  \gamma \frac{\|\bm{\widehat{A}_{c}} \Delta \phi(\bm{x}) ^{(c)}\|^{2}}{1-\epsilon_{2}}, \|\Delta \phi(\bm{x}) ^{(c,c')}\|^{2} > \frac{\|\bm{\widehat{A}_{c}} \Delta \phi(\bm{x}) ^{(c,c')}\|^{2}}{1+\epsilon_{1}}\Bigg) = \\
    \Pro_{}\Bigg(\|\Delta \phi(\bm{x}) ^{(c)}\|^{2} <  \frac{\|\bm{\widehat{A}_{c}} \Delta \phi(\bm{x}) ^{(c)}\|^{2}}{1-\epsilon_{2}}, \|\Delta \phi(\bm{x}) ^{(c,c')}\|^{2} >  \frac{\|\bm{\widehat{A}_{c}} \Delta \phi(\bm{x}) ^{(c,c')}\|^{2}}{1+\epsilon_{1}}\Bigg) \geq \\
     1 - \Pro_{}\Bigg(\frac{\|\bm{\widehat{A}_{c}} \Delta \phi(\bm{x}) ^{(c,c')}\|^{2}}{\|\Delta \phi(\bm{x}) ^{(c,c')}\|^{2}} > {1+\epsilon_{1}}\Bigg) - \Pro_{}\Bigg(\frac{\|\bm{\widehat{A}_{c}} \Delta \phi(\bm{x}) ^{(c)}\|^{2}}{\|\Delta \phi(\bm{x}) ^{(c)}\|^{2}} < {1-\epsilon_{2}}\Bigg)
     \\ = F_{{C}}^{\chi}(1+\epsilon_{1}) - F_{C}^{\chi}(1-\epsilon_{2}) \text{ where $F_{{C}}^{\chi}(u) = $ cdf of a $\chi^{2}_{(C-1)}$ random variable at $(C-1)u$.}
\end{multline}
In order to show how the above inequality is obtained, consider two random variables $X$ and $Y$. Then:
\[\Pro_{}(X<x,Y>y) = 1 - \{\Pro_{}(X>x,Y>y) + \Pro_{}(X>x,Y<y)\} - \Pro_{}(X<x,Y<y)\]
\[= 1 - \Pro_{}(X>x) - \Pro_{}(X<x,Y<y) \geq 1 - \Pro_{}(X>x) - \Pro_{}(Y<y)\]
In the above steps, the first equality follows from the law of total probability, the second equality is obtained by marginalizing out $Y$ and the third inequality follows from the fact that $\Pro_{}(Y<y) = \Pro_{}(X<x,Y<y) + \Pro_{}(X>x,Y<y) \geq \Pro_{}(X<x,Y<y) \Rightarrow \Pro_{}(X<x,Y<y) \leq \Pro_{}(Y<y)$. 

Also:
\begin{equation}
    \label{eq:32}
    \frac{\|\bm{\widehat{A}_{c}} \Delta \phi(\bm{x}) ^{(c,c')}\|^{2}}{(1+\epsilon_{1})} > \gamma \frac{\|\bm{\widehat{A}_{c}} \Delta \phi(\bm{x}) ^{(c)}\|^{2}}{(1-\epsilon_{2})} \Rightarrow \frac{\|\bm{{A}_{c}} \Delta \phi(\bm{x}) ^{(c,c')}\|^{2}}{(1+\epsilon_{1})} > \gamma \frac{\|\bm{{A}_{c}} \Delta \phi(\bm{x}) ^{(c)}\|^{2}}{(1-\epsilon_{2})}
    %\\
    %\text{ut } {\|\bm{{A}_{c}} \Delta \phi(\bm{x}) ^{(c,c')}\|^{2}} > \frac{{(1+\epsilon_{1})}}{(1-\epsilon_{2})}{\|\bm{{A}_{c}} \Delta \phi(\bm{x}) ^{(c)}\|^{2}} \text{ w.p. }\geq b_{A}\Big(\frac{{1+\epsilon_{1}}}{1-\epsilon_{2}}\Big). 
\end{equation}
\[\text{But from \textbf{Theorem 3} - } {\|\bm{{A}_{c}} \Delta \phi(\bm{x}) ^{(c,c')}\|^{2}} > \gamma \frac{{(1+\epsilon_{1})}}{(1-\epsilon_{2})}{\|\bm{{A}_{c}} \Delta \phi(\bm{x}) ^{(c)}\|^{2}} \text{ w.p. }\geq b_{A}\Big(\gamma \frac{{1+\epsilon_{1}}}{1-\epsilon_{2}},L\Big).\]
Note that the derivation of (\ref{eq:30}) is completely oblivious of $\Delta \phi(\bm{x}) ^{(c)}$ and $\Delta \phi(\bm{x}) ^{(c,c')}$, and depends only on the distribution of $\bm{{A}_{c}}$. And the derivation of (\ref{eq:32}) is completely oblivious of the distribution of $\bm{{A}_{c}}$.
%Therefore, using (\ref{eq:30}), (\ref{eq:31}) and (\ref{eq:32}), we get:
Leveraging these two facts, we use (\ref{eq:30}) and (\ref{eq:32}) to get:
%Finally, from (\ref{eq:30}), (\ref{eq:31}) and (\ref{eq:32}), we get:
\begin{multline}
    \label{eq:33}
    \|\Delta \phi(\bm{x}) ^{(c,c')}\|^{2} >  \frac{\|\bm{\widehat{A}_{c}} \Delta \phi(\bm{x}) ^{(c,c')}\|^{2}}{1+\epsilon_{1}} > \gamma \frac{\|\bm{\widehat{A}_{c}} \Delta \phi(\bm{x}) ^{(c)}\|^{2}}{1-\epsilon_{2}} > \gamma \|\Delta \phi(\bm{x}) ^{(c)}\|^{2}
    \\
    \text{ w.p. } \geq b_{A}\Big(\gamma \frac{{1+\epsilon_{1}}}{1-\epsilon_{2}},L\Big)(F_{{C}}^{\chi}(1+\epsilon_{1}) - F_{C}^{\chi}(1-\epsilon_{2})).
\end{multline}

Notice that (\ref{eq:33}) holds for all $\epsilon_{1},\epsilon_{2} \in (0,1)$. Thus, taking max over all $\epsilon_{1},\epsilon_{2} \in (0,1)$, we get:
\begin{equation}
    \label{eq:34}
    \Pro_{}\Big(\|\Delta \phi(\bm{x}) ^{(c,c')}\|^{2} > \gamma \|\Delta\phi(\bm{x}) ^{(c)}\|^{2}\Big) \geq \max_{0 < \epsilon_{1},\epsilon_{2} < 1} b_{A}\Big(\gamma\frac{{1+\epsilon_{1}}}{1-\epsilon_{2}},L\Big)(F_{{C}}^{\chi}(1+\epsilon_{1}) - F_{C}^{\chi}(1-\epsilon_{2})).
\end{equation}
This proves the first part of \textbf{Theorem 4}.

%The second part of \textbf{Theorem 4} is simply the application of a bound on the value of $\Pro_{}\Big(\frac{\|\bm{\widehat{A}_{c}} \Delta \phi(\bm{x}) ^{(c,c')}\|^{2}}{\|\Delta \phi(\bm{x}) ^{(c,c')}\|^{2}} > {1+\epsilon_{1}}\Big)$ and that of $\Pro_{}\Big(\frac{\|\bm{\widehat{A}_{c}} \Delta \phi(\bm{x}) ^{(c)}\|^{2}}{\|\Delta \phi(\bm{x}) ^{(c)}\|^{2}} < {1-\epsilon_{2}}\Big)$

The second part of \textbf{Theorem 4} involves simply upper-bounding $\Pro_{}\Big(\frac{\|\bm{\widehat{A}_{c}} \Delta \phi(\bm{x}) ^{(c,c')}\|^{2}}{\|\Delta \phi(\bm{x}) ^{(c,c')}\|^{2}} > {1+\epsilon_{1}}\Big)$ and $\Pro_{}\Big(\frac{\|\bm{\widehat{A}_{c}} \Delta \phi(\bm{x}) ^{(c)}\|^{2}}{\|\Delta \phi(\bm{x}) ^{(c)}\|^{2}} < {1-\epsilon_{2}}\Big)$. Essentially, we derive some upper bounds on the values of $\Pro_{}(\chi^{2}_{(C-1)} > (1+\epsilon)(C-1))$ and $\Pro_{}(\chi^{2}_{(C-1)} < (1-\epsilon)(C-1))$.

We provide tighter bounds than the ones provided in Lemma 1.3 of \cite{JL_Gaussian4}
by using a result provided in Lemma 1, Page 1325 of \cite{laurent2000adaptive} which is as follows:
\begin{equation}
    \label{eq:35}
    \Pro_{}(\chi^{2}_{k} \geq k + 2\sqrt{kx}+2x) \leq e^{-x} \text{ and } \Pro_{}(\chi^{2}_{k} \leq k - 2\sqrt{kx}) \leq e^{-x}
\end{equation}
Now substituting $(1+\epsilon)k = k + 2\sqrt{kx} + 2x$ gives us $x = \frac{k}{2}(1+\epsilon - \sqrt{1+2\epsilon})$. Similarly, substituting $(1-\epsilon)k = k - 2\sqrt{kx}$ gives us $x = k\epsilon^{2}/4$.
Using this, we get:
\begin{equation}
    \label{eq:36}
    \Pro_{}(\chi^{2}_{k} \geq (1+\epsilon)k) \leq \exp(-\frac{k}{2}(1+\epsilon - \sqrt{1+2\epsilon})) \text{ and } \Pro_{}(\chi^{2}_{k} \leq (1-\epsilon)k) \leq \exp(-\frac{k}{4}\epsilon^{2}).
\end{equation}

Substituting the bounds obtained above in (\ref{eq:36}) along with $k = (C-1)$ in (\ref{eq:30}) and then repeating the rest of the process (of the proof) up to (\ref{eq:34}), we obtain the second part of \textbf{Theorem 4}.

This concludes the proof of \textbf{Theorem 4}.

\subsection{Proof of Theorem 5}
\label{sec:3_8}
Finally, we restate and then prove the last theorem, i.e. \textbf{Theorem 5}. 
\medskip{}

\begin{theorem20*}
Let there be $N_{c}$ examples belonging to class $c \in \bm{[C]}$. Under the assumption of \textbf{Theorem 1}, define $\kappa_{c}^{*} \triangleq{} \min_{c' \ne c}\kappa_{c',c}$ ($\kappa_{c',c}$'s are the same as in \textbf{Theorem 2}). 
Then the expected number of examples in class $c$ correctly classified, say $N_{c}^{\text{corr}}$, is: %(the lower bound holds regardless of the validity of the assumption):
\small
\[N_{c}^{\text{corr}} = N_{c}\Big(1 - \exp\Big(-\Big(\frac{\Gamma(\beta+1)\log(1+\kappa_{c}^{*})}{L}\Big)^{1/\beta}\Big)\Big) \geq N_{c}\Big(1 - \exp\Big(-\Big(\frac{\Gamma(\beta+1)\log(2)}{L}\Big)^{1/\beta}\Big)\Big)\]
\normalsize
The lower bound provided above holds regardless of the validity of the assumption in \textbf{Theorem 1}.
\end{theorem20*}
\medskip{}

\textbf{Proof}: Consider a point $\bm{x}$ belonging to class $c$. Let the probability of it belonging to $c$, predicted by the network be $\widehat{y}$. As per the notation used in \textbf{Theorem 2}, the probability of $\bm{x}$ belonging to $c' \ne c$, predicted by the network, is $(1-\widehat{y})/\kappa_{c',c}$ and under the assumption of \textbf{Theorem 1}, $\kappa_{c',c}$ is a constant for all points belonging to $c$. %Regardless of whether the assumption of \textbf{Theorem 1} holds or not, 

Now, for $\bm{x}$ to be correctly classified by the network, we must have $\widehat{y} > (1-\widehat{y})/\kappa_{c',c}$ $\forall$ $c' \ne c$. In other words, we must have $\widehat{y} > (1-\widehat{y})/\kappa_{c}^{*}$ where $\kappa_{c}^{*} = \min_{c' \ne c}\kappa_{c',c}$. This is equivalent to $\widehat{y} > 1/(\kappa_{c}^{*}+1)$ or $(-\log(\widehat{y}))^{1/\beta} < (\log(\kappa_{c}^{*}+1))^{1/\beta}$. 
Now using (3) of the main paper and substituting the value of $\mu$ obtained from \textbf{Lemma 1} in the main paper, we have:
\begin{equation}
    \label{eq:37}
     \Pro_{}((-\log(\widehat{y}))^{1/\beta} < (\log(\kappa_{c}^{*}+1))^{1/\beta}) = 1 - \exp\Big(-\Big(\frac{\Gamma(\beta+1)\log(1+\kappa_{c}^{*})}{L}\Big)^{1/\beta}\Big) = p_{\text{acc}}(L,\kappa_{c}^{*}) 
\end{equation}

Using (\ref{eq:37}), the expected number of correctly classified examples actually belonging to $c$, $N_{c}^{\text{corr}}$ will be:
\begin{equation}
    \label{eq:38}
     N_{c}^{\text{corr}} = %N_{c}\Bigg(1 - \exp\Bigg(-\Big(\frac{\Gamma(\beta+1)\log(1+\kappa_{c}^{*})}{L}\Big)^{1/\beta}\Bigg)\Bigg)
     N_{c}p_{\text{acc}}(L,\kappa_{c}^{*})
\end{equation}
This proves the first part of \textbf{Theorem 5}. 

We have $\kappa_{c}^{*} \geq 1$ since $\kappa_{c',c} \geq 1$ $\forall$ $c' \ne c$. It can be verified that $p_{\text{acc}}(L,\kappa_{c}^{*})$ is an increasing function of $\kappa_{c}^{*}$. Therefore:
\begin{equation}
    \label{eq:39}
     N_{c}^{\text{corr}} \geq
     N_{c}p_{\text{acc}}(L,1) = N_{c}\Big(1 - \exp\Big(-\Big(\frac{\Gamma(\beta+1)\log(2)}{L}\Big)^{1/\beta}\Big)\Big)
\end{equation}
Note that (\ref{eq:39}) can be also derived by just computing the probability with which $\widehat{y} > (1-\widehat{y})$ (notice that this is equivalent to setting $\kappa_{c}^{*} = 1$) which always ensures that $\bm{x}$ is correctly classified. In other words, the probability with which correct classification occurs is at least as large as the probability with which $\widehat{y} > (1-\widehat{y})$. %Thus (\ref{eq:39}) holds regardless of whether the assumption of \textbf{Theorem 1} is valid or not.
Thus (\ref{eq:39}) holds irrespective of the validity of the assumption of \textbf{Theorem 1}.

This completes the proof of \textbf{Theorem 5}.

\subsection{Additional Experiments}
\label{sec:3_5}
%\textbf{FILL IT!}
We firstly describe the two synthetic datasets (SYN-1 and SYN-2) that we talked about in the main paper. {We also provide code to generate SYN-1 and SYN-2.}

\textbf{First synthetic dataset (SYN-1)}: Here, we considered $10$ dimensional random Gaussian vectors as input belonging to one out of $20$ possible classes. Denote one such random input by $\bm{z} = [z_{1},z_{2},\ldots,z_{10}]^{T}$. The class of $\bm{z}$ is determined by the range in which the value of $h(\bm{z}) = \sum_{i=1}^{10}\exp(z_{i})/10$ lies in. Specifically, we have interval points $\{r_{1},r_{2},\ldots,r_{19}\}$. Then, $z$ belongs to class number $0$ if $h(z) < r_{1}$, $z$ belongs to class number $i$ ($0 < i < 19$) if $r_{i} \leq h(z) < r_{i+1}$ and $z$ belongs to class number $19$ if $h(z) \geq r_{19}$. The interval points were chosen such that the number of training points in each interval are all nearly the same. For SYN-1, we fitted a fully connected neural network (the entire architecture can be found in Subsection \ref{sec:3_6}) consisting of three hidden layers with ReLU activation. The value of $n$ used was $80$. 
The total number of training set and test set examples were 16000 and 4000 (randomly chosen), respectively.
%The final loss value was $L = 0.5632$. 
We trained the model for 100 epochs and the obtained loss value over the test set (after 100 epochs) was $L = 0.5632$. 
We observed that $\beta = 2$ results in the closest resembling exponential distribution for SYN-1. 
%\medskip{}

\textbf{Second synthetic dataset (SYN-2)}: This is similar to the previous dataset, except that the function for the classification rule was taken to be %$h(\bm{z}) = \sum_{i=1}^{10}z_{i}^{4}/10$.
a simple linear function, $h(\bm{z}) = (\sum_{i=1}^{5}2z_{i} + \sum_{i=6}^{10}z_{i})/5$.
We used the same architecture as that used for SYN-1 (and so $n = 80$). 
%Once again, the value of $n$ was chosen to be $80$. 
Here, the total number of training set and test set examples were 24000 and 6000 (again randomly chosen), respectively.
%The final loss value was $L = 0.1889$ in this case. 
The test set loss value after training the model for 20 epochs was $L = 0.1889$.
%Also, in
In this case, $\beta = 1.4$ results in the closest resembling exponential distribution. 

%\Cref{fig:1_c} and \Cref{fig:1_d} 
Figure 4c and Figure 4d 
in the main paper show the distribution of $\Big(-\log(\widehat{y_{i}}^{(c(\bm{x_{i}}))})\Big)^{1/\beta}$ along with the closest fit exponential distribution for SYN-1 with $\beta = 2$ and SYN-2 with 
$\beta = 1.4$, respectively, over the test set. 

Also, \Cref{tab_syn1} and \Cref{tab_syn2} show the probability of the inter-class distance being more than the intra-class distance for SYN-1 and SYN-2 respectively, over the test set. %Here, $p_{1}$ and $p_{2}$ are as defined in the main paper.

\begin{table}[!h]
\scalebox{0.9}{
\centering
   \subfloat[SYN-1 \label{tab_syn1}]{
     \centering
     \begin{tabular}{|l|l|l|l|}
        \hline
                       Class 1 ($c_{1}$) & Class 2 ($c_{2}$) & $p_{1}$ & $p_{2}$     \\
                       \hline
                       %\rule{0pt}{10pt}
        0 & 1 & 0.7575 & 0.8448 \\
        \hline
        1 & 2 & 0.8317 & 0.7751 \\
        \hline
        5 & 6 & 0.6196 & 0.6319 \\
        \hline
        16 & 17 & 0.6577 & 0.5257 \\
        \hline
        11 & 13 & 0.8278 & 0.7645 \\
        \hline
        7 & 9 & 0.8216 & 0.7836 \\
        \hline
        12 & 15 & 0.9675 & 0.8782 \\
        \hline
        14 & 18 & 0.9951 & 0.9251 \\
        \hline
        5 & 10 & 0.9985 & 0.9931 \\
        \hline
        3 & 12 & 1.0000 & 0.9964 \\
        \hline
    \end{tabular}
   }
   \hspace{0.05 in}
   \subfloat[SYN-2 \label{tab_syn2}]{
     \centering
     \begin{tabular}{|l|l|l|l|}
        \hline
                       Class 1 ($c_{1}$) & Class 2 ($c_{2}$) & $p_{1}$ & $p_{2}$     \\
                       \hline
        0 & 1 & 0.8169 & 0.9583 \\
        \hline
        1 & 2 & 0.8927 & 0.9153 \\
        \hline
        5 & 6 & 0.9123 & 0.9445 \\
        \hline
        16 & 17 & 0.9036 & 0.8848 \\
        \hline
        11 & 13 & 0.9998 & 0.9995 \\
        \hline
        7 & 9 & 0.9990 & 0.9999 \\
        \hline
        12 & 15 & 1.0000 & 1.0000 \\
        \hline
        14 & 18 & 1.0000 & 1.0000 \\
        \hline
        5 & 10 & 1.0000 & 1.0000 \\
        \hline
        3 & 12 & 1.0000 & 1.0000   \\
        \hline
    \end{tabular}
   }
}
\caption{Sample probabilities of inter-class distance being more than the intra-class distance for (a) SYN-1 with $L = 0.5632$ and (b) SYN-2 with $L = 0.1889$. $p_{1}$ and $p_{2}$ are as defined in the text.}
\label{tab_syn}
\end{table}

Observe that in both \Cref{tab_syn1} and \Cref{tab_syn2}, the values of $p_{1}$ and $p_{2}$ are lower when $|c_{1}-c_{2}|$ is small as compared to when $|c_{1}-c_{2}|$ is large. This is because, based on the description of the two datasets, if $|c_{1}-c_{2}| < |c_{1}-c_{3}|$ then it implies that the interval corresponding to $c_{2}$ is closer to the interval corresponding to $c_{1}$ than that corresponding to $c_{3}$, leading to poorer separability between the  points
%examples
belonging to $c_{1}$ and $c_{2}$ as compared to the points 
%examples 
belonging to $c_{1}$ and $c_{3}$.

In \Cref{tab_syn1} with $L = 0.5632$, observe that $p_{1}$ and $p_{2}$ values for the first 4 entries of the table where $|c_{1}-c_{2}| = 1$ is in the range of $(0.6,0.85)$ while their values for the last 2 entries of the table where $|c_{1}-c_{2}| \geq 5$ is nearly 1. %The corresponding loss value ($L$) is $0.5632$.
In order to illustrate the effect of the choice of $\kappa_{c,c'}$ other than $(C-1)$, we report the values of $b_{c}(0.5632)$ with $\beta = 2$, $C = 20$, $\kappa_{c,c'} = 0.2(C-1)$ %(which indicates more similarity/poorer separation between $c$ and $c'$ than if we set $\kappa_{c,c'} = C-1$) 
and $\kappa_{c,c'} = 5(C-1)$ %(which indicates lesser similarity/better separation between $c$ and $c'$ than if we set $\kappa_{c,c'} = C-1$) 
(which indicate greater and lesser similarity or poorer and better separation, respectively, between $c$ and $c'$, than if we would have set $\kappa_{c,c'} = C-1$).
%and
These turn out to be approximately $0.5494$ and $0.7295$, respectively.

In \Cref{tab_syn2} with $L = 0.1889$, observe that $p_{1}$ and $p_{2}$ values for the first 4 entries of the table where $|c_{1}-c_{2}| = 1$ is in the range of $(0.8,0.95)$ while their values for the last 4 entries of the table where $|c_{1}-c_{2}| \geq 3$ is exactly 1.
Here, the values of $b_{c}(0.1889)$ with $\beta = 1.4$, $C = 20$, $\kappa_{c,c'} = 0.2(C-1)$ %(which indicates more similarity/poorer separation between $c$ and $c'$ than if we set $\kappa_{c,c'} = C-1$)
and $\kappa_{c,c'} = 5(C-1)$ %(which indicates lesser similarity/better separation between $c$ and $c'$ than if we set $\kappa_{c,c'} = C-1$) 
turn out to be approximately $0.7749$ and $0.8986$, respectively.

Thus, even here, the obtained values are consistent with the observed values.

%\subsection{Architectures used in the experiments}
\subsection{Architectures used and some more training details}
\label{sec:3_6}
\begin{figure}[!h]
\centering 
\subfloat[CIFAR]{
	\includegraphics[width = 43.65 mm, height = 108.15 mm]{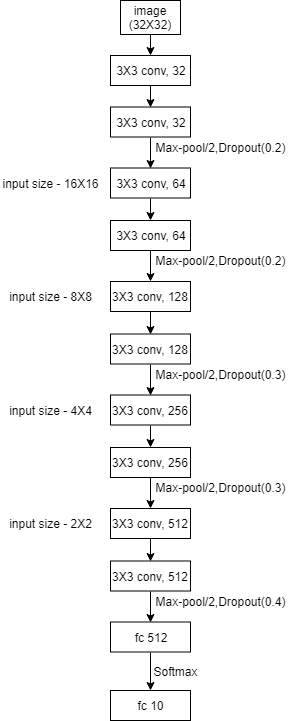}
	\label{fig:arch_cifar}
}
\hspace{5 mm}
\subfloat[MNIST]{
	\includegraphics[width = 43.65 mm, height = 54.3 mm]{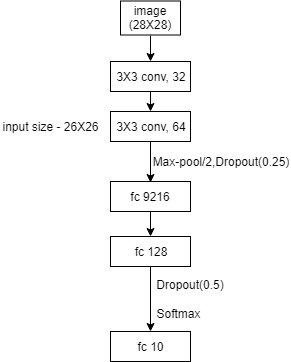}
	\label{fig:arch_mnist}
}
\hspace{5 mm}
\subfloat[SYN-1]{
	\includegraphics[width = 12.15 mm, height = 42.15 mm]{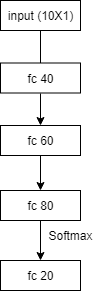}
	\label{fig:arch_syn}
}
\caption{Block-diagram of architectures used in our experiments for (a) CIFAR-10, (b) MNIST and (c) SYN-1 (as well as SYN-2). Unless otherwise specified, the default activation function is ReLU everywhere. Also, `fc' denotes a fully connected layer.}
\label{fig:arch}
\end{figure}

Finally, we show the block-diagrams of the network architectures used for the 4 datasets in \Cref{fig:arch}. Observe that the values of $n$ in \Cref{fig:arch_cifar}, \Cref{fig:arch_mnist} and \Cref{fig:arch_syn} are $512, 128$ and $80$, respectively. As usual, the last layer in all 3 cases is a fully connected Softmax layer.

\begin{comment}
\begin{figure}[!h]
\centering 
\subfloat[CIFAR]{
	\includegraphics[width = 29.1 mm, height = 72.1 mm]{CIFAR_arch.png}
	\label{fig:arch_cifar}
}
\hspace{5 mm}
\subfloat[MNIST]{
	\includegraphics[width = 29.1 mm, height = 36.2 mm]{MNIST_arch.png}
	\label{fig:arch_mnist}
}
\hspace{5 mm}
\subfloat[SYN]{
	\includegraphics[width = 8.1 mm, height = 28.1 mm]{SYN_arch.png}
	\label{fig:arch_syn}
}
\caption{}
\label{fig:arch}
\end{figure}
\end{comment}

%All the network fitting experiments (done on the four datasets) have been performed using Keras.
The models described earlier were fitted %on their respective datasets 
with Keras (using NVIDIA GeForce 940MX GPU). The batch size used for CIFAR-10, MNIST, SYN-1 and SYN-2 were 64, 128, 200 and 400, respectively. The training set and test set of MNIST and CIFAR-10 were used as provided with sizes of the training set being 60000 and 50000, respectively. As mentioned in the main paper also, size of the test set was 10000 for both the datasets. 
%\\
%\\
%\\
%\\
%\\
%\\
%\\
%\\
%\\
%\\
%\bibliographystyle{plain}%siamplain}
%\bibliography{references}

\end{document}